%% file: aistats_main.tex
\documentclass[twoside]{article}

\usepackage[accepted]{aistats2024}

\usepackage{hyperref}       % hyperlinks
\usepackage{url}            % simple URL typesetting
\usepackage{booktabs}       % professional-quality tables
\usepackage{amsfonts}       % blackboard math symbols
\usepackage{nicefrac}       % compact symbols for 1/2, etc.
\usepackage{microtype}      % microtypography
\usepackage{xcolor}         % colors

\usepackage{tikz}
\usetikzlibrary{positioning}
\usepackage{cleveref}

\usepackage[english]{babel}
\usepackage[round]{natbib}

\usepackage{amsmath}
\usepackage{graphicx}

\input{math_commands.tex}

\usepackage{hyperref}
\usepackage{url}
\usepackage{algorithm}
\usepackage{algpseudocode}
\usepackage{multirow}
\usepackage{neuralnetwork}
\usepackage{subcaption}
\newtheorem{theorem}{Theorem}
\newtheorem{remark}{Remark}

\newenvironment{proof}{\paragraph{Proof:}}{\hfill$\square$}

\begin{document}

% If your paper is accepted and the title of your paper is very long,
% the style will print as headings an error message. Use the following
% command to supply a shorter title of your paper so that it can be
% used as headings.
%
%\runningtitle{I use this title instead because the last one was very long}

% If your paper is accepted and the number of authors is large, the
% style will print as headings an error message. Use the following
% command to supply a shorter version of the authors names so that
% they can be used as headings (for example, use only the surnames)
%
%\runningauthor{Surname 1, Surname 2, Surname 3, ...., Surname n}

\twocolumn[

\aistatstitle{On The Temporal Domain of Differential Equation Inspired Graph Neural Networks}

\aistatsauthor{ Moshe Eliasof$^1$ \And Eldad Haber$^2$ \And  Eran Treister$^3$ \And Carola-Bibiane Sch{\"o}nlieb$^1$ }

\aistatsaddress{ $^1$ University of Cambridge \And  $^2$ University of British Columbia \And $^3$ Ben-Gurion University of the Negev } ]

\begin{abstract}
Graph Neural Networks (GNNs) have demonstrated remarkable success in modeling complex relationships in graph-structured data. A recent innovation in this field is the family of Differential Equation-Inspired Graph Neural Networks (DE-GNNs), which leverage principles from continuous dynamical systems to model information flow on graphs with built-in properties such as feature smoothing or preservation. However, existing DE-GNNs rely on first or second-order temporal dependencies. In this paper, we propose a neural extension to those pre-defined temporal dependencies. We show that our model, called TDE-GNN, can capture a wide range of temporal dynamics that go beyond typical first or second-order methods, and provide use cases where existing temporal models are challenged. We demonstrate the benefit of learning the temporal dependencies using our method rather than using pre-defined temporal dynamics on several graph benchmarks.
\end{abstract}

\section{Introduction}

Graph neural networks (GNNs) are now ubiquitous in diverse applications from
social media to chemistry and physical systems, see \cite{wu2020comprehensive,wang2021review} and references within. In recent years,  it has been shown that GNNs can be viewed as dynamical systems. Specifically, Ordinary Differential Equations (ODE) based methods have been found to be useful, providing understandable behavior, such as smoothing \citep{poli2019graph, chamberlain2021grand}, energy conservation \citep{eliasof2021pde, rusch2022graph}, anti-symmetry \citep{gravina2023antisymmetric}, pattern formation \citep{wang2022acmp, choi2022gread}, and more. We refer to this family of architectures as DE-GNNs. 

Many works in the field of DE-GNNs consider {\em stationary data}, that is, data that is not time dependent. As such, most works focus on the \emph{spatial} interactions between nodes, and how to better model them. At the same time, the majority of existing DE-GNNs employ first-order temporal dynamics, which, as we show later in Example \ref{ex:nonlinear_pendulum}, can be limiting. 
Therefore, in this paper, we study the importance of the time domain of DE-GNNs, and propose a novel mechanism to model the temporal order and dependencies of the underlying ODEs of GNN layers in a data-driven fashion.
As we show, the utilization of the proposed advanced temporal domain learning mechanism offers a practical performance advantage, while also naturally bridging between works that have been proposed to handle tasks for {\em non-stationary data} by
time-dependent graph neural networks (TD-GNNs), as in 
\cite{dyngrae_1,guan2022dynagraph, xiong2020dynamic, pilva2022learning, longa2023graph}.

\emph{The goal of this work is to develop and study a novel mechanism to model the time domain of DE-GNNs.} Our approach, called \emph{TDE-GNN}\footnote{Read as Teddy-GNN.}
, is based on learning (i) the \emph{temporal order}, and, (ii) the \emph{temporal dependency} in a data-driven fashion. The temporal order defines the order of the underlying dynamics as favored by the data, while the temporal dependency specifies the relationship between intermediate DE steps. 
To the best of our knowledge, this is the first work to study the temporal domain of DE-GNNs, in the sense that it learns higher-order DEs in a general manner. All DE-GNNs known to us, utilize either first or second-order time dependencies.
In other words,  existing DE-GNNs assume fixed temporal behavior, which is constant in time and is pre-defined. 
This shortcoming, as we show later, can be rather limiting when complex phenomena are to be modeled. 
Our TDE-GNN can also be viewed as an extension for Residual Networks \citep{he2016deep}, which is aimed to incorporate node features from previous time steps (i.e., layers) in a learnable manner.

\section{Related Work}
\label{sec:related}
\textbf{Graph Neural Networks Inspired by Differential Equations.}
Adopting the interpretation of convolutional neural networks (CNNs) as discretizations of ODEs and PDEs  \citep{RuthottoHaber2018, chen2018deep, zhang2019linked} to GNNs, works like GCDE \citep{poli2019graph}, GODE \citep{zhuang2020ordinary}, GRAND \citep{chamberlain2021grand}, PDE-GCN\textsubscript{D} \citep{eliasof2021pde}, GRAND++ \citep{thorpe2022grand} and others, propose to view GNN layers as time steps in the integration of the non-linear heat equation. This perspective allows to control the diffusion (smoothing) in the network, to understand oversmoothing \citep{nt2019revisiting,oono2020graph,cai2020note} in GNNs. Thus, works like \cite{chien2021adaptive, luan2022revisiting, di2022graphGRAFF} propose to utilize a \emph{learnable} diffusion term, thereby alleviating oversmoothing. Other architectures like PDE-GCN\textsubscript{M} \citep{eliasof2021pde} and GraphCON \citep{rusch2022graph} propose to mix diffusion and oscillatory processes (e.g., based on the wave equation) to avoid oversmoothing by introducing a feature energy preservation mechanism. Nonetheless, as noted in \cite{rusch2023survey}, besides alleviating oversmoothing, it is also important to design GNN architectures with improved expressiveness. Recent examples of such networks are \cite{gravina2023antisymmetric} that propose an anti-symmetric GNN to alleviate over-squashing \citep{alon2021oversquashing}, \cite{wang2022acmp, choi2022gread} that formulate a reaction-diffusion GNN to enable non-trivial pattern growth, \cite{convectionGNN} that propose a convection-diffusion based GNN, advection-reaction-diffusion to allow directed information transportation \citep{eliasof2023adr}, and \cite{maskey2023fractional} that formalize a fractional Laplacian ODE based GNN with improved expressiveness. 
A common theme of most of the aforementioned works, is the focus on the \emph{spatial} term of the ODE, while the temporal term is set to be of first or second order. In this work, we propose to extend the family of ODE-inspired GNNs from the perspective of the \emph{temporal} domain.

\textbf{The Temporal Domain in Graph Neural Networks.} 
In recent years, GNNs for spatio-temporal data were developed. Some examples are \cite{gclstm,gconvlstm, tgcn} that combine graph convolution with LSTM mechanisms, and other combines graph attention with temporal mechanisms, as in \cite{a3tgcn}. Other works like \cite{evolvegcn,bai2020adaptive} propose adaptive graph convolutions for temporal graphs. It has also been shown in \cite{gutteridge2023drew} that adjacency matrix update according to intermediate node features is useful for long-range benchmarks. Furthermore, recent works have shown that GNNs for temporal graph datasets can benefit from the interpretation and construction of ordinary differential equations. For example, it was shown in \cite{diffusionTemporal,RDtemporal} that reaction and diffusion systems can improve traffic prediction, and it was shown in \cite{choi2023climate} that advection and diffusion can improve weather forecasting performance. However, all the considered works discussed here utilize first-order temporal dynamics, while focusing on the spatial term of the ordinary differential equation. In this paper, we explore and study the temporal domain in the context of DE-GNNs, and show its importance to model complex systems and improve performance.

\section{Mathematical Background and Motivation}
\label{sec:math}
In this section, we provide a related mathematical overview, and motivate the necessity of our TDE-GNN architecture that enables higher-order DE-GNNs, by an example where first-order models are challenged.

\textbf{Notations.}
We consider a graph ${G=(V,E)}$, where $V$ is a set of
 $n$ nodes, and $E \subseteq V \times V$ is a set of $m$ edges.
The $i$-th node is associated with a possibly time-dependent hidden feature vector 
 $f_i(t) \in \mathbb{R}^k$. Let $F(t) = [f_0(t), \ldots, f_{n-1}(t)]^\top$ be a $n \times k$  matrix that represents the node state (features) at time $t$. 

\paragraph{Differential Equations Inspired GNNs (DE-GNNs).}
The basic idea of the DE-GNN family of architectures is to discretize the following ODE:
\begin{subequations}
\label{eq:basicODE}
\begin{align}
\label{eq:basicODE_a}
&{\frac {\partial F}{\partial t}} = s\left(F(t); G\right) \\
\label{eq:basicODE_b}
&F(t=0) = \bfF^{(0)}\end{align}
\end{subequations}

where $s(F(t); G)$ is a spatial operator that depends on the graph $G$ and the node features $F(t)$. Specifically, it is common to employ graph diffusion,  combined with a channel mixing operator implemented by a multilayer perceptron (MLP). Some examples of such methods were proposed in \cite{chamberlain2021grand, eliasof2021pde, thorpe2022grand, choi2022gread}, 
and others. Because we focus on the \emph{temporal} component of the ODE in this paper, we employ a similar spatial term that combines diffusion and channel mixing, as discussed later. Then, the graph ODE in \Cref{eq:basicODE} is discretized in time, until time $T$, typically with the forward Euler method. 
The chosen discretization times are considered as GNN layers, with a total of $L$ time steps with step size $h$ such that $T=hL$.

For stationary problems (e.g., node classification), the input consists of a single time step at time $T_0$. 
Given input features $\bfI^{(0)} \in \mathbb{R}^{n \times k_{in}}$, we embed them using an MLP to obtain the initial conditions of the ODE, denoted by $\bfF^{(0)}\in \mathbb{R}^{n \times k}$. For spatio-temporal tasks (e.g., forecasting node quantities), the node features $[\bfI^{(0)},\ldots,\bfI^{(r)}]$ are provided at sampled times $[T_0, \ldots, T_r]$, and embedded in latent space. The node features at the final GNN layer $\bfF^{(L)}$ are then fed to a classifier to output the desired shaped prediction to be compared with the labeled data, depending on the task. 

The right-hand side of \Cref{eq:basicODE_a} describes the \emph{spatial} behavior of the DE-GNN, and has been thoroughly studied in previous works, as discussed in Section \ref{sec:related}. The left-hand side, which describes the temporal order and dynamics of the ODE, however, did not receive significant attention, to the best of our knowledge. Most of the works known to us, also discussed in Section \ref{sec:related}, consider only first-order time dynamics, with the exception of \cite{eliasof2021pde, rusch2022graph} that limit their models to second-order dynamics. 
Thus, in this work, we focus on the left-hand side of \Cref{eq:basicODE_a}, which describes the temporal \emph{order} and \emph{dynamics} of DE-GNNs.  We will show that learning the temporal domain of the DE-GNN offers two major benefits: (i) interpretable learned weights in the time domain, and (ii) improved downstream task performance.

\textbf{Problem Formulation.} 
The downstream tasks considered in this work aim to predict node values, either by regression or classification. The common theme among the considered tasks is that we view them as the prediction of the time and space evolution of the node features, given past and current node features. A popular approach 
is to treat the problem by combining a GNN with time series mechanisms such as LSTM \citep{hochreiter1997long} or GRU \citep{GRU}, to predict the future state by using the current state, as discussed in Section \ref{sec:related}.
While such techniques have shown promising results, we provide Example \ref{ex:nonlinear_pendulum},  where a standard GNN-LSTM is challenged, in the sense that it does not perform better than a naive solution. We attribute this shortcoming to the basic assumption of models like LSTM, that future predictions can be based on the previous state, effectively assuming first-order dynamics, which may not be sufficient to model higher-order phenomena, as shown below.  

\begin{example} {(Nonlinear Pendulum)}{
\label{ex:nonlinear_pendulum}
\em 
Let us consider the problem of a nonlinear pendulum. The
pendulum can be treated as a graph with two nodes. The first node $v_0$ is fixed
at $(0,0)$, and the second, $v_1$, is located at  $(x_1(t), y_1(t))$ that evolve in time. We illustrate the pendulum system in Figure \ref{fig:pendulumIllustration}.
\begin{figure}[]
    \centering    \includegraphics[width=0.8\linewidth]{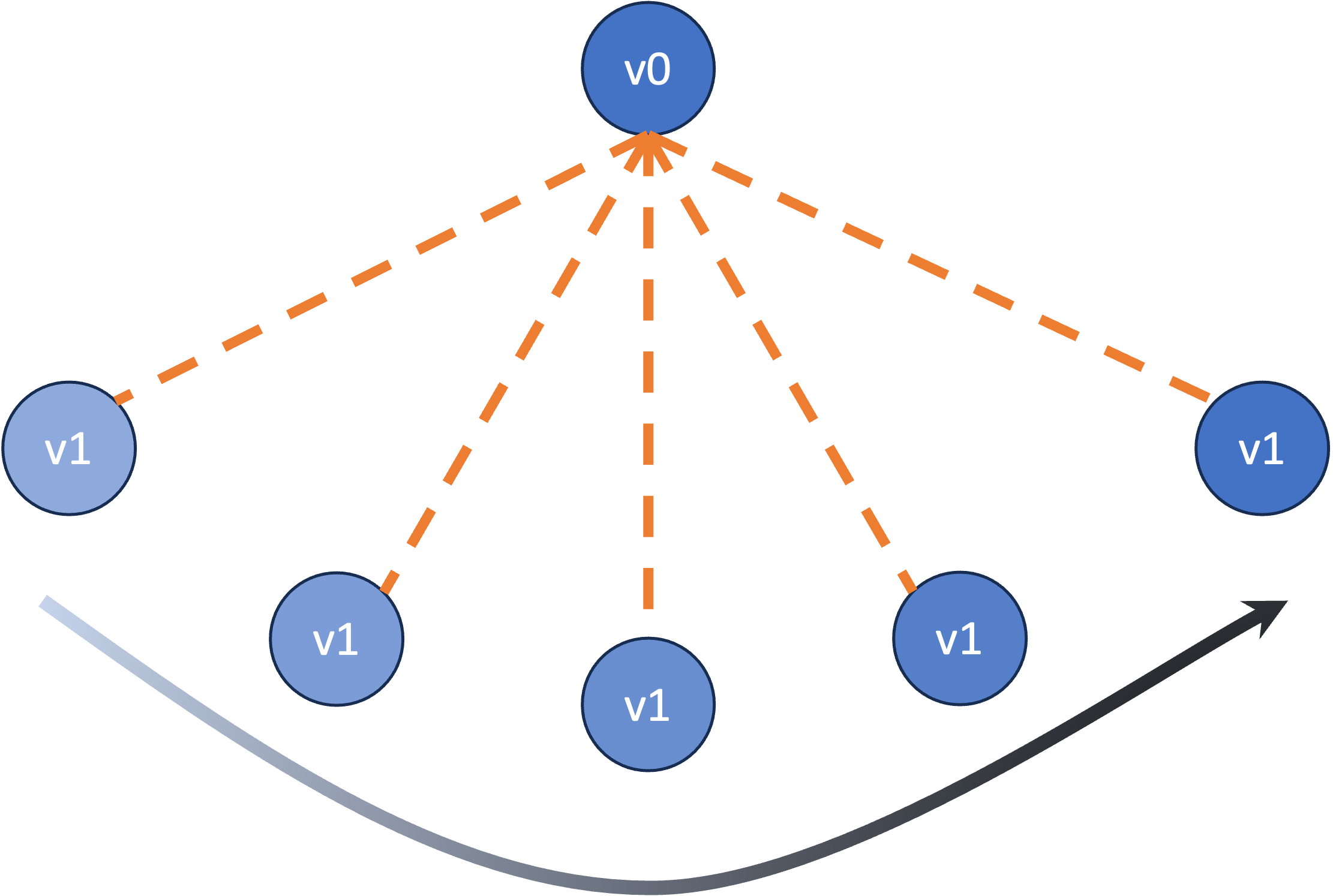} 
    \caption{An illustration of a pendulum.}
\label{fig:pendulumIllustration}
\end{figure}
Evaluating the coordinates of the nodes $v_0, v_1$
at time $t$ can be done by solving the Newtonian mechanics that define the nonlinear pendulum's motion. Specifically, it is described by the following equation:
\begin{equation}
    \label{eq:pendulumEquation}
    \frac{\partial^2 F }{\partial t^2} = q(F),
\end{equation}
where $q(F)$ is a gradient of the energy that characterizes the behavior of the pendulum discussed in Appendix \ref{app:pendulum}. 

We discretize \Cref{eq:pendulumEquation} using the leapfrog method \citep{ascher2008numerical}, to generate
a time series data of the pendulum vertices locations. Recall that node $v_0$ is static, and remains in (0,0), while $v_1$ moves according to \Cref{eq:pendulumEquation}.  We plot the location of $v_1$ in Figure~\ref{fig:pendulum_data}.
\begin{figure}[t]
    \centering
    \begin{subfigure}[t]{.8\linewidth}
\includegraphics[width=1\linewidth]{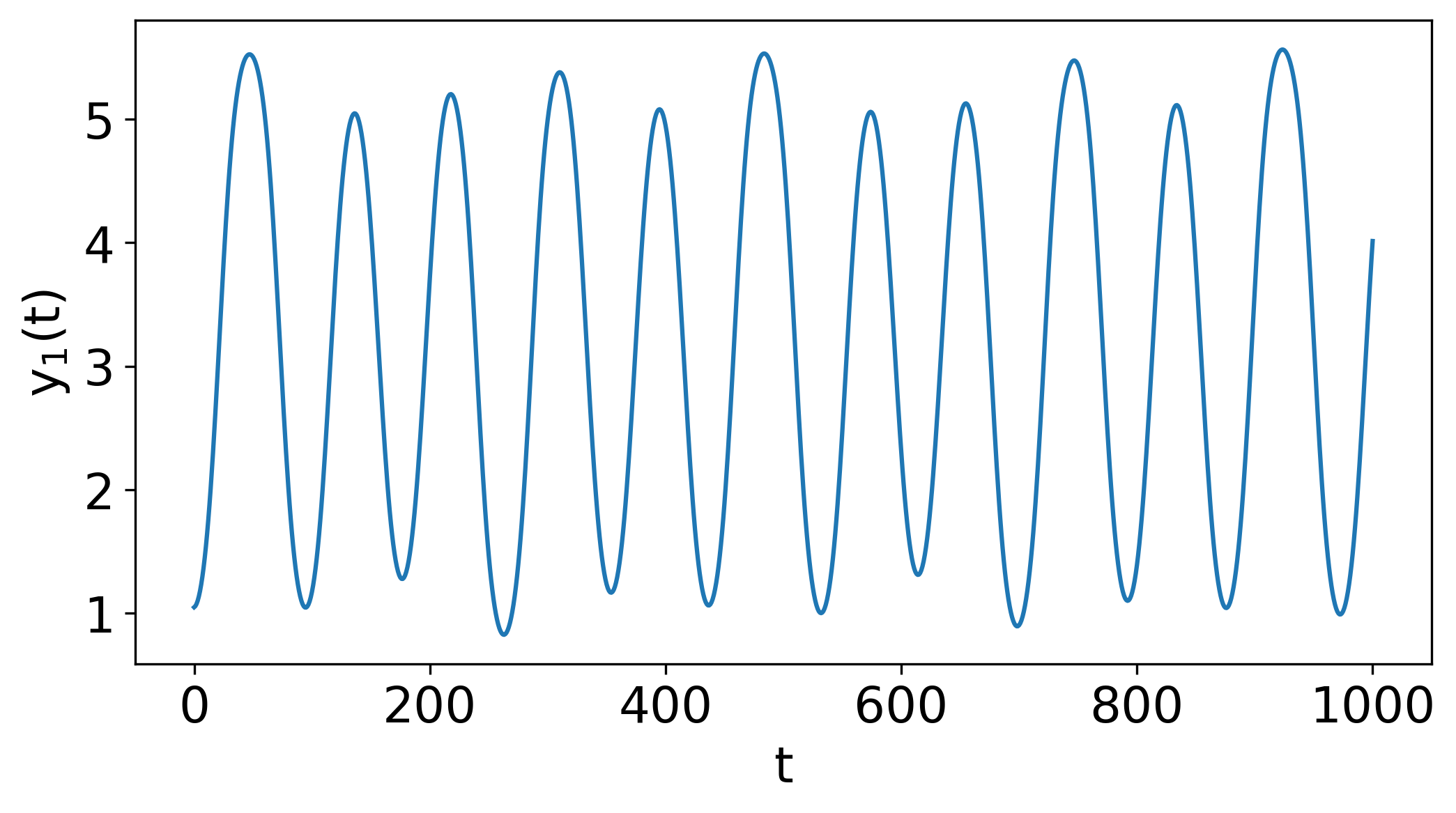}
                    \caption{}
        \label{fig:pendulum_data}
\end{subfigure} 
    \begin{subfigure}[t]{.8\linewidth}
    \includegraphics[width=1\linewidth]{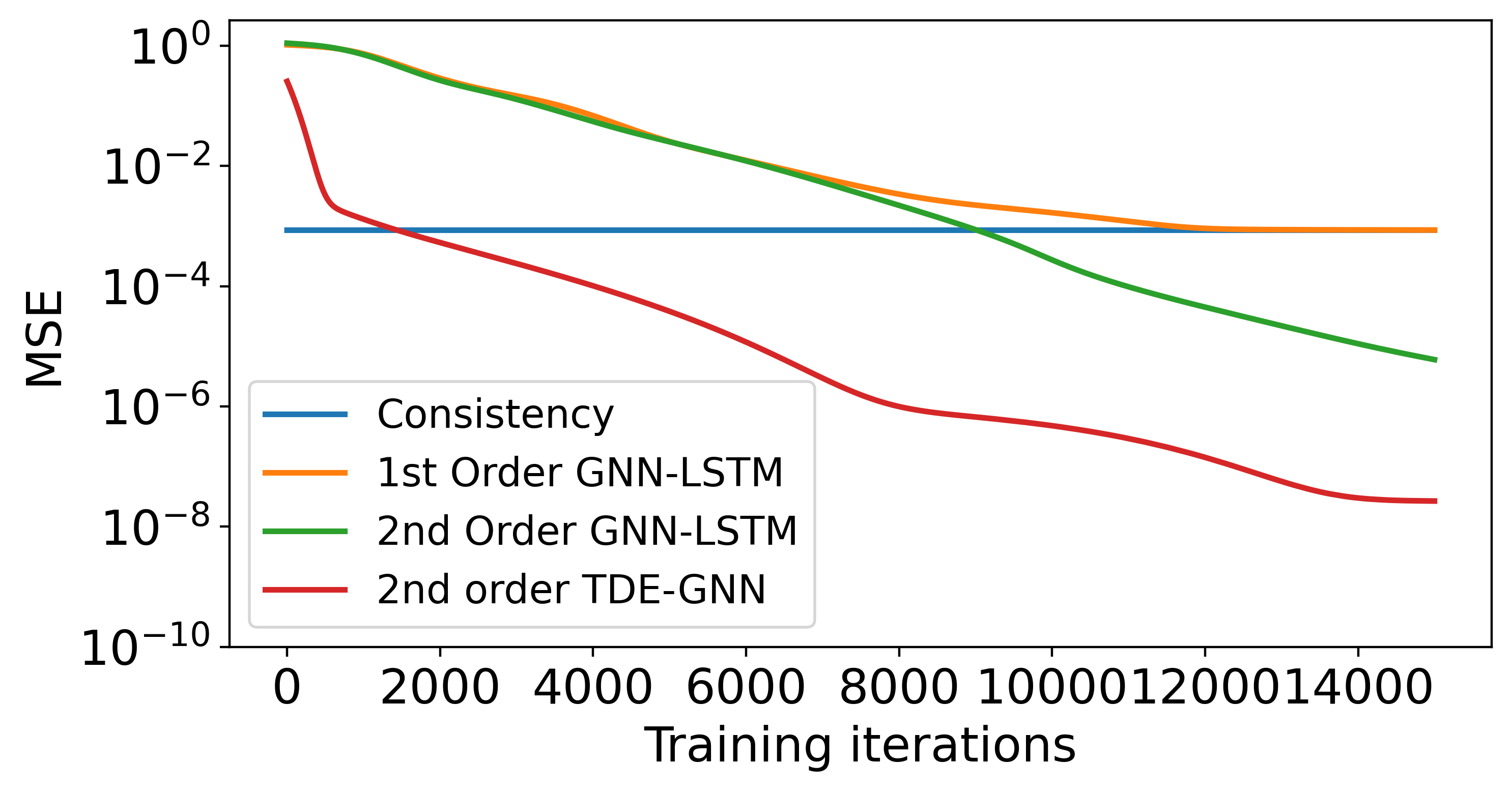}
                        \caption{}
        \label{fig:pendulum_results}
    \end{subfigure}

    \caption{The pendulum location prediction example. (a). Pendulum $y_1(t)$ coordinate vs. time (b) The prediction performance of naive, 1st, and 2nd order models. Higher-order models offer improved predictions.}
    \label{fig:pendulum}
\end{figure}

We define a task whose inputs are observed locations of the pendulum nodes, and the goal is to predict the future locations of nodes.
We consider four possible prediction models to address this task: (i) A \emph{naive} prediction model, oftentimes called a \emph{consistency} model, that simply outputs the latest available state. Formally, it is given by  $\bfF^{(l+1)} = \bfF^{(l)}$. (ii) A GNN-LSTM, similar to \cite{gconvlstm}, (iii) a second order GNN-LSTM discussed in Appendix \ref{app:pendulum}, and, (iv) our TDE-GNN limited to second order for a fair comparison, to be defined later in Section \ref{sec:method}.

We report the obtained prediction mean squared error (MSE) compared to the ground-truth data in Figure \ref{fig:pendulum_results}. We observe that the first-order GNN-LSTM model performs as good as the naive model of consistency, thereby not offering improved results as one would like. 
To understand the limitations of GNN-LSTM, it is key to recall \Cref{eq:pendulumEquation}, and see that a pendulum's motion involves a second-order system. That is, to predict a future location $\bfF^{(l+1)}$, one is required to use both $\bfF^{(l)}$ and $\bfF^{(l-1)}$. However, the first-order GNN-LSTM mechanism considers only the latest state (node features) $\bfF^{(l)}$. Indeed, when considering a second-order GNN-LSTM that involves both $\bfF^{(l)}$ and $\bfF^{(l-1)}$
one can obtain improved performance. Finally, we see that our TDE-GNN limited to second-order offers further prediction performance improvement.
The convergence curves of the considered networks are plotted in Figure~\ref{fig:pendulum_results}.

This example demonstrates that the order (length) of the history used for prediction is important. If too short of a history is used, it may be impossible to accurately predict the behavior of systems with order higher than the available history length.
Since for many problems, the order is unknown and can vary in time, we allow TDE-GNN to learn the order from the data.} 
\end{example}

\section{Time Dependent Differential Equations Inspired GNNs}
\label{sec:method}

Example \ref{ex:nonlinear_pendulum} demonstrates the importance of utilizing higher-order temporal behavior to fit complex data. Motivated by this example, we propose and study, a method that
can learn the \emph{temporal} domain of the DE-GNNs from
the data, in addition to leveraging useful spatial terms, as proposed in other works and discussed in Section \ref{sec:related}. Therefore, we call our method \emph{TDE-GNN}. 

\subsection{TDE-GNN Learns Higher-Order DEs}
\label{sec:TDEGNN}

As discussed, previous works have so far mostly 
considered first order time dependent GNNs as in \Cref{eq:basicODE}, whose forward Euler discretization reads:
\begin{equation}
    \label{eq:TimeDiscExistingMethods}
    \bfF^{(l+1)} = \bfF^{(l)} + hs( \bfF^{(l)}; G),
\end{equation}
where $\bfF^{(l)} \in \mathbb{R}^{n\times k}$ are the node features at the $l$-th layer, $h$ is a positive discretization step size, and $s$ is the spatial term. To focus on the proposed temporal mechanism, in this work we follow previous works that combine graph diffusion with channel mixing. We elaborate on the implementation of $s$ in Appendix \ref{app:implementingSpace}.

In this paper, we generalize and study the time order and dynamics, and introduce a TDE-GNN layer that stems from the following ODE, with a maximal order of $o \geq 1$, which is a hyperparameter:
\begin{equation}
    \label{eq:contADRms}
   \sum_{p=1}^{o} c_p{\frac {\partial^p F}{\partial t^p}}=  s\left( F(t); G \right),
\end{equation}
accompanied by the initial conditions 
\begin{equation}
\label{eq:initialConditions}
    {\frac {\partial^{p}F}{\partial t^{p}}}\Big|_{t=0} = F^{(p)}(t=0) \quad p=0,\ldots, o-1 .
\end{equation}
The forward Euler discretization of \Cref{eq:contADRms} yields a layer of our extended, higher-order, TDE-GNN layer: 
\begin{equation}
\label{eq:TDE_TimeDisc}
\bfF^{(l+1)} = \sum_{p=1}^{o} c_p(\mathcal{H}_o^{(l)}) \bfF^{(l-p+1)} + hs(\bfF^{(l)}; G).
\end{equation}
Here, we define $\bfc(\mathcal{H}_{o}^{(l)}) = [c_1(\mathcal{H}_{o}^{(l)}),\ldots, c_{o}(\mathcal{H}_{o}^{(l)})] \in \mathbb{R}^{o} $, which are learned weights based on previous node features of up to order $o$, formally denoted by:
\begin{equation}
\label{eq:history}
\mathcal{H}_o^{(l)} = [\bfF^{(l)} \| \bfF^{(l-1)} \| \ldots \| \bfF^{(l-o+1)}] \in \mathbb{R}^{o \times n \times k},
\end{equation} where $\|$ denotes the stacking operation.
 It is important to note that \Cref{eq:TDE_TimeDisc} models the dynamics of the $l$-th layer with $o-1$ previous layers, and therefore yields a discretization of an ODE of order $o$. For stability of computation, and for interoperability, we demand that
$\sum_{p=1}^{o} c_p = 1$. This constraint ensures that the coefficients approximate derivatives of
the data up to $o$-th order \citep{EvansPDE}. 
Because in this paper we focus on the temporal behavior of the underlying graph ODE, in Section \ref{sec:implementingTime} we describe our mechanism that learns the temporal order and dynamics encoded by $\bfc(\mathcal{H}_{o}^{(l)})$, and for completeness, in Appendix \ref{app:implementingSpace} we discuss the spatial term of our TDE-GNN. 

\textbf{Understanding the learned coefficients $\bfc(\mathcal{H}_{o}^{(l)})$.}  Before we proceed with implementation details, it is important to note that \Cref{eq:TDE_TimeDisc} extends the idea of residual networks. Typical residual networks (as in ResNet \citep{he2016deep}) can be obtained by \Cref{eq:TDE_TimeDisc} with $o=1$ and $c_1=1$, which yields the forward Euler method that is often used to discretize diffusive GNNs \citep{chamberlain2021grand, eliasof2021pde}. Also, second order oscillatory GNNs as proposed in \cite{eliasof2021pde, rusch2022graph}, can be implemented by \Cref{eq:TDE_TimeDisc}
 with $o=2$ and $\bfc = [c_1,c_2] = [2, -1]$. 
Overall, our TDE-GNN can implement, as well as extend, both of these types of architectures by learning higher-order dynamics with adaptive coefficients $\bfc(\mathcal{H}_{o}^{(l)})$.
Those extensions allow our TDE-GNN to model a diverse family of dynamics that cannot be obtained with the aforementioned methods. We now provide Example \ref{ex:thirdOrder} that shows how a third-order DE-GNN is implemented by our method and how the learned coefficients can be interpreted.

\setcounter{example}{1}
\begin{example} {(3rd Order TDE-GNN)}{
\label{ex:thirdOrder} \em 
We now draw a link between a third-order TDE-GNN (with an order hyperparameter $o=3$), and a third-order ODE. Note that every set of coefficients $\{c_1,c_2,c_3 \}$ that sum to $1$, with a step size $h=1$
can be spanned by the basis:
\begin{equation}  
\label{eq:3rdOrder}
 \begin{pmatrix}
    c_1 \\ c_2 \\ c_3
\end{pmatrix} =
\alpha_1
\begin{pmatrix}
    1 \\ 0 \\ 0
\end{pmatrix} +
\alpha_2 
\begin{pmatrix}
    2 \\ -1 \\ 0
\end{pmatrix} +
\alpha_3
\begin{pmatrix}
    2 \\ -2 \\ 1
\end{pmatrix},
\end{equation}
with the constraint $\sum_{i=1}^{3} \alpha_i=1$. Note that the vectors that multiply $\alpha_1,\alpha_2$ and $\alpha_3$ correspond to a first-order, second-order, and third-order finite difference, respectively (i.e., $\frac{\partial F(t)}{\partial t}$, $\frac{\partial^2 F(t)}{\partial t^2}$, and $\frac{\partial^3 F(t)}{\partial t^3}$).
Since the basis in \Cref{eq:3rdOrder}
is complete, for any $c_1,c_2,c_3$ that sum to 1, there exists a 3rd order differential equation whose discretization yields the same coefficients.}
\end{example}

Thus, treating the temporal domain in  DE-GNNs using our learnable framework 
allows us to {\em reveal and understand the order of the underlying time-dependent process in a data-driven fashion}. Furthermore, as we have shown in Example \ref{ex:nonlinear_pendulum}, such a treatment can be crucial to accurately model data that stems from higher-order phenomena. Namely, if the system we intend to predict is of order $o$ and we do not expose the network to a history of at least $o$ time steps, 
then accurate predictions may not be possible. 

\subsection{Implementing $\bfc(\mathcal{H}_{o}^{(l)})$}
\label{sec:implementingTime}

At the core of our TDE-GNN stands the learning of the temporal coefficients $\bfc(\mathcal{H}_{o}^{(l)})$, with two key requirements for a valid implementation: (i) the vector $\bfc$ sums to 1, i.e., $\sum_{p=1}^{o}c_p(\mathcal{H}_{o}^{(l)}) = 1$, and, (ii) the entries of $\bfc(\mathcal{H}_{o}^{(l)})$ can be any real-valued number. These requirements offer both training stability (due to the normalization in requirement (i)), and the approximation of a finite order derivative \citep{EvansPDE}. We now discuss two implementations that we consider in this paper, and later, in our experiments in Section \ref{sec:experiments}, we compare their performance.

\textbf{Direct parameterization.} Perhaps the most intuitive implementation is obtained by \emph{direct} parameterization, where we directly learn a vector $\tilde{\bfc} \in \mathbb{R}^{o}$, and divide it by the sum of its entries (to satisfy requirement (i)), leading to the temporal coefficients vector:
\begin{equation}
    \label{eq:directParam}
    \bfc = \frac{\tilde{\bfc}}{\sum_{p=1}^{o}\tilde{c}_p}.
\end{equation}
Note that in this case, the coefficients vector $\bfc$ is not directly influenced by the history $\mathcal{H}_o^{(l)}$, but it is still optimized according to the history via backpropagation.

\textbf{Attention-based parameterization.} In addition to the direct parameterization, we propose a novel mechanism that leverages an attention mechanism as in \cite{vaswani2017attention}. A key feature of the attention mechanism is that it outputs a pairwise score of its input. The \emph{novelty} here is to apply the attention mechanism on the \emph{temporal} dimension. To this end, by collecting and appropriately shaping the node features of the previous $o$ layers, denoted by $\mathcal{H}_{o}^{(l)}\in \mathbb{R}^{o \times n \times k}$ as defined in  \Cref{eq:TDE_TimeDisc}, and feeding it to an attention layer \citep{vaswani2017attention}, one obtains a pairwise score map $\mathcal{S} \in [0,1]^{o \times o}$. The last row in $\mathcal{S}$ represents the temporal scores of the $l$-th layer with the previous layers $o-1$. Clearly, requirement (i) is met by the SoftMax function used in the attention mechanism in \cite{vaswani2017attention}. However, a SoftMax function yields non-negative pairwise values, which do not satisfy requirement (ii). Such a limitation will prevent, for instance, the ability to implement the oscillatory equation discussed in \ref{sec:TDEGNN} using our TDE-GNN. Therefore, we follow the same implementation as in \cite{vaswani2017attention} up to the SoftMax step. Instead, we only apply a normalization of the obtained pairwise interaction map by dividing it by its sum. This procedure satisfies both requirements (i) and (ii). We provide further details about the implementation in Appendix  \ref{app:implementingTemporal}.

\begin{figure}[t]
    \centering
    \begin{tabular}{c}
    \includegraphics[width=0.8\linewidth]{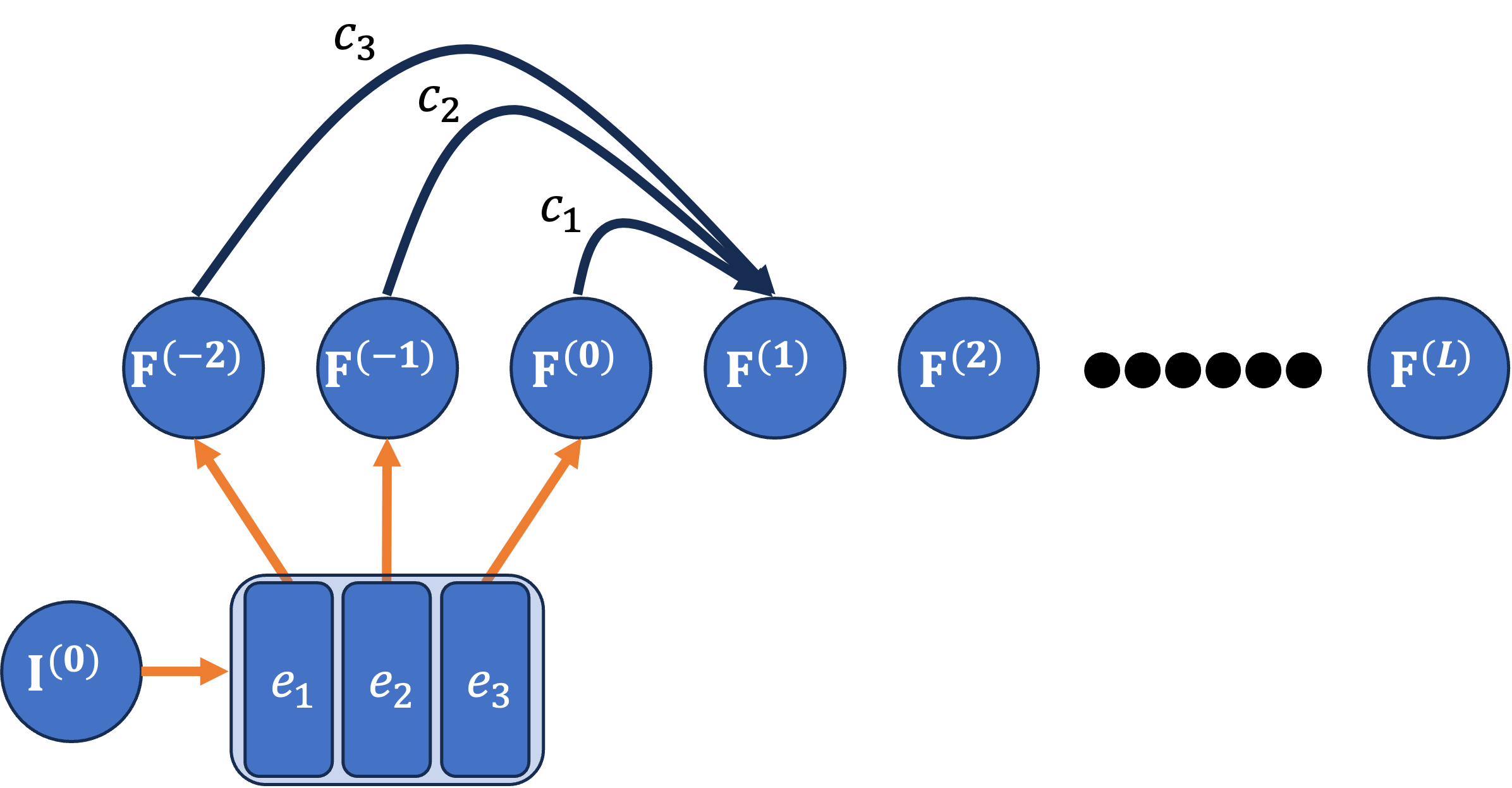} 
    \end{tabular}
    \caption{The embedding of input features $\bfI$ using an MLPs ($e_1,e_2,e_3$) to obtain $o=3$ initial conditions, followed by our TDE-GNN for stationary problems.}
    \label{fig:static}
\end{figure}

\textbf{Initial conditions.} When considering high-order ODEs, the aspect of the initial conditions of the model is important \citep{AscherPetzoldODEs}. 
We consider two use cases that are treated differently. First, when solving a stationary problem such as node classification, where only a single initial temporal condition is available, we use $o$ MLPs to embed this single state into $o$ states, 
and then use the network in \Cref{eq:TDE_TimeDisc}. This initialization, as well as the application of a TDE-GNN at the first layer, is illustrated in Figure \ref{fig:static}.
\iffalse
\begin{figure}[h]
    \centering
\begin{tikzpicture}

\draw[red, line width=0.5mm] (0.0, 0.0) circle [radius=0.5] node (a) {Input};
\draw[blue, line width=0.5mm] (-1.6, 1.6) circle [radius=0.45] node (b) {$F_{-2}$};
\draw[blue, line width=0.5mm] (0.0, 1.6) circle [radius=0.45] node (c) {$F_{-1}$};
\draw[blue, line width=0.5mm] (1.6, 1.6) circle [radius=0.45] node (d) {$F_{0}$};
\draw[black, line width=0.5mm] (3.2, 1.6) circle [radius=0.45] node (e) {$F_{1}$};
\draw[black, line width=0.5mm] (3.9, 1.6) circle [radius=0.02] node (g) {};
\draw[black, line width=0.5mm] (4.05, 1.6) circle [radius=0.02] node (h) {};
\draw[black, line width=0.5mm] (4.2, 1.6) circle [radius=0.02] node (j) {};

\draw[black, line width=0.5mm] (5.0, 1.6) circle [radius=0.45] node (f) {$F_{L}$};

\draw[red, line width=0.5mm] (5.0, 0.0) circle [radius=0.55] node (k) {Output};

\draw[->, line width=0.5mm] (a) -- (b);
\draw[->, line width=0.5mm] (a) -- (c);
\draw[->, line width=0.5mm] (a) -- (d);
\draw[->, line width=0.5mm] (f) -- (k);

\draw[->, line width=0.5mm] (b.north) to [out=60, in=1200] (e.north);
\draw[->, line width=0.5mm] (c.north) to [out=60, in=1200] (e.north);
\draw[->, line width=0.5mm] (d.north) to [out=60, in=1200] (e.north);

 \end{tikzpicture}
    \caption{Initialization of the network for stationary data for a neural order of $o=3$. }
    \label{fig-net1}
\end{figure}
\fi

\begin{figure}[t]
    \centering    \includegraphics[width=0.8\linewidth]{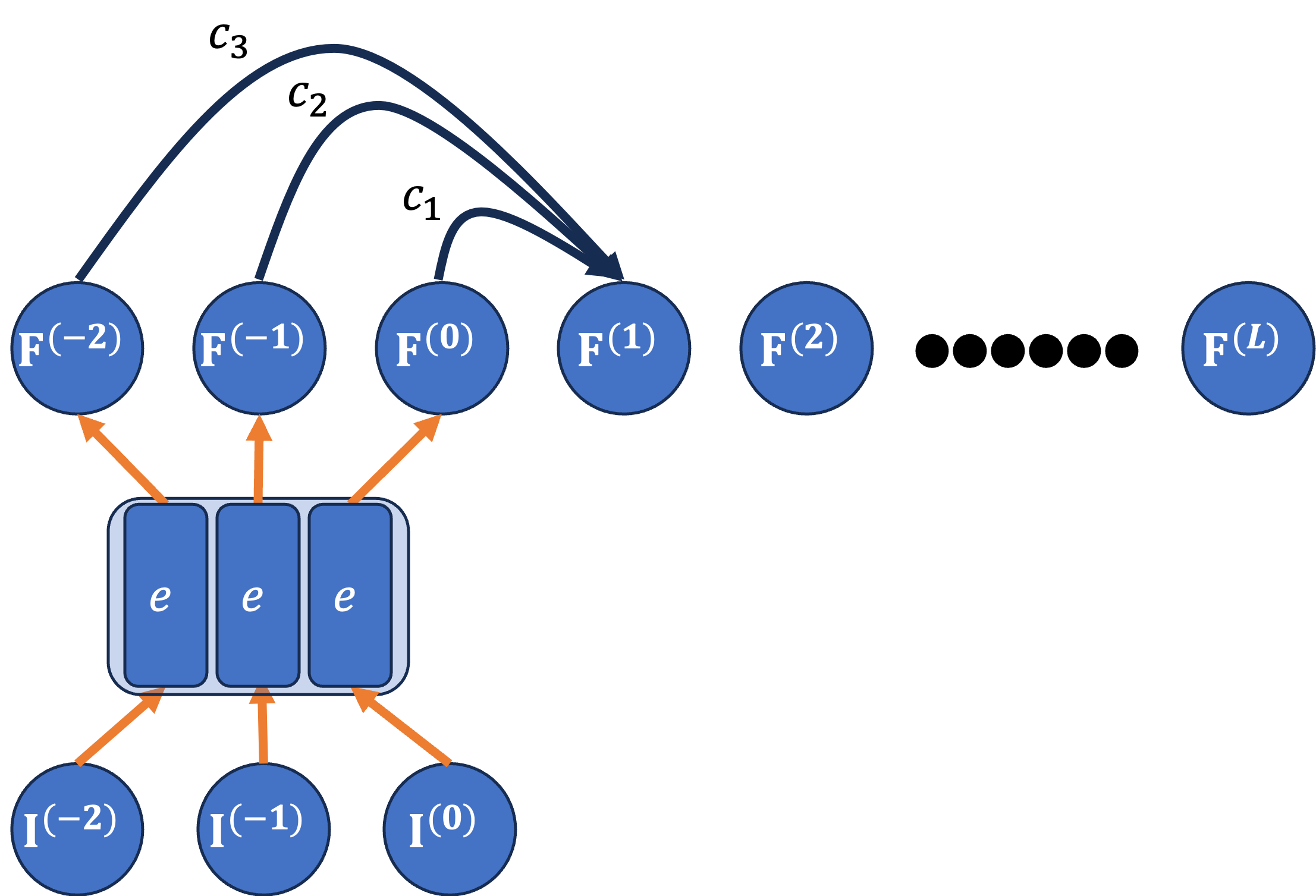} 
    \caption{The initialization of TDE-GNN for spatio-temporal data with a history of $o=3$.}
\label{fig:temporal}
\end{figure}

For time series graph problems where we have a time series as input, 
we use at least $o$ historical data in order to initialize the states.
In this case, the frequency of the {\em observed input} can be different than the frequency of the hidden space $\bfF^{(l)}$ that discretizes the ODE, in the sense that we can use more hidden layers than observed inputs. Upon receiving at least $o$ input observations, we embed them using an MLP to obtain $o$ hidden initial conditions. In Figure \ref{fig:temporal}, we illustrate the described process as well as the application of a TDE-GNN layer to the inputs. Past the initialization step, in both stationary and non-stationary cases, the features update relies on previously computed hidden node features, as described in \Cref{eq:TDE_TimeDisc}.

\textbf{Complexity.} Compared to existing DE-GNN methods, our TDE-GNN involves additional $o-1$ additions of previous node features, to allow modeling differential equations of order $o$, and achieve improved performance, as we show in Section \ref{sec:experiments}. If the coefficients $\bfc$ are obtained using the \emph{direct} parameterization, then $o-1$ scalar multiplications are required. If the \emph{attention} based mechanism is utilized to learn and evaluate $\bfc$, adding $\mathcal{O}(n \cdot k \cdot o^2)$ multiplications. We note that  $o$, the order hyperparameter of TDE-GNN,  is typically significantly smaller than the number of channels $k$ and nodes $n$, because it is bounded by the number of layers $L$. In Appendix \ref{app:runtimes} we report the training and inference runtimes of the proposed implementations.

\textbf{Properties of TDE-GNN.}
\label{sec:properties}
Our TDE-GNN draws inspiration from a stable discrete process of ODE integration, that generates future time values, which is regarded as the node features evolution throughout the layers. Therefore, a natural question that arises is whether the obtained network is stable. Indeed, if the proposed architecture is unstable, then it may be difficult to fit the data, or, the network may not generalize well (see \cite{haber2017stable} for stability definition and a thorough discussion). To this end, we prove the following theorem in Appendix  \ref{app:derivations}. \\
\begin{theorem}(Stability of TDE-GNN).
\label{thm:stability}
 For the discretization of \Cref{eq:TDE_TimeDisc}, there exists a vector $\bfc = [c_1, \ldots, c_{o}]$ such that the discrete solution is stable.
\end{theorem}
In our ablation study in Section \ref{exp:ablation}, we verify Theorem \ref{thm:stability}, and show that the learnable weights $\bfc$ can be interpreted as finite difference derivatives.

\section{Experiments}
\label{sec:experiments}

\begin{table*}[h]
  \begin{center}
  
     \footnotesize
   \setlength{\tabcolsep}{3pt}
  \begin{tabular}{lcccccc}
    \toprule
    Dataset & Squirrel & Film &  Chameleon &  Citeseer & Pubmed & Cora \\
    Homophily & 0.22 & 0.22 & 0.23 & 0.71 & 0.74 & 0.81 \\
        \midrule
    \textbf{General GNNs} \\
    $\,$GCN  & 23.96$\pm$2.01 & 26.86$\pm$1.10 &  28.18$\pm$2.24& 73.68$
    \pm$1.36 & 88.13$\pm$0.50 & 85.77$\pm$1.27  \\
    $\,$GAT & 30.03$\pm$1.55 & 28.45$\pm$0.89 & 42.93$\pm$2.50 &
   74.32$\pm$1.23 & 87.62$\pm$1.10 & 86.37$\pm$0.48 \\
    $\,$GCNII\textsuperscript{$*$} & 38.47$\pm$1.58 & 32.87$\pm$1.30 &   60.61$\pm$3.04 & 77.13$\pm$1.48 & 90.30$\pm$0.43 & 88.49$\pm$1.25 \\
    $\,$Geom-GCN\textsuperscript{$*$}& 38.32$\pm$0.92 & 31.63$\pm$1.15 &  60.90$\pm$2.81 & 77.99$\pm$1.15 & 90.05$\pm$0.47 & 85.27$\pm$1.57  \\
    $\,$NSD\textsuperscript{$*$}  &  56.34$\pm$1.32 & {37.79$\pm$1.15}  & 68.68$\pm$1.58 & 77.14$\pm$1.57 & 89.49$\pm$0.40 & 87.14$\pm$1.13  \\
    $\,$GGCN  & 55.17$\pm$1.58 & {37.81$\pm$1.56} &  71.14$\pm$1.84  & 77.14$\pm$1.45 & 89.15$\pm$0.37 & 87.95$\pm$1.05 \\
    $\,$H2GCN  & 36.48$\pm$1.86 & 35.70$\pm$1.00 & 60.11$\pm$1.71 & 77.11$\pm$1.57 & 89.49$\pm$0.38 & 87.87$\pm$1.20  \\ 
    $\,$FAGCN & 42.59$\pm$0.69 & 34.87$\pm$1.35 & 55.22$\pm$2.11 & 74.01$\pm$1.85 & 76.57$\pm$1.88 & 86.34$\pm$0.67  \\
    $\,$GPRGNN & 31.61$\pm$1.24 & 34.63$\pm$1.22 & 46.58$\pm$1.71 & 77.13$\pm$1.67 & 87.54$\pm$0.38 & 87.95$\pm$1.18 \\
    $\,$GRAFF\textsuperscript{$*$} &  59.01$\pm$1.31 & 37.11$\pm$1.08 & 71.38$\pm$1.47 & 77.30$\pm$1.85 & 90.04$\pm$0.41 & 88.01$\pm$1.03\\
    $\,$LINKX  & 61.81$\pm$1.80 &  36.10$\pm$1.55 & 68.42$\pm$1.38 & 73.19$\pm$0.99 & 87.86$\pm$0.77 & 84.64$\pm$1.13\\
    $\,$ACMII\textsuperscript{$*$} & 67.40$\pm$2.21 & 37.09$\pm$1.32 & {74.76$\pm$2.20}& 77.12$\pm$1.58 & 89.71$\pm$0.48 & 88.25$\pm$0.96 \\
    \midrule
    \textbf{GNNs Inspired by DEs} \\
    $\,$GRAND & 40.05$\pm$1.50 & 35.62$\pm$1.01 &  54.67$\pm$2.54 & 76.46$\pm$1.77 & 89.02$\pm$0.51 & 87.36$\pm$0.96\\
    $\,$PDE-GCN\textsuperscript{$*$} & N/A & N/A &   66.01$\pm$2.11 & 78.45$\pm$1.98 & 89.93$\pm$0.62 & 88.60$\pm$1.77 \\
    $\,$GRAND++ & 40.06$\pm$1.70 & 33.63$\pm$0.48 &  56.20$\pm$2.15 & 76.57$\pm$1.46  & 88.50$\pm$0.35 &88.15$\pm$1.22   \\ 
    $\,$GREAD\textsuperscript{$*$} & 59.22$\pm$1.44 & 37.90$\pm$1.17 & 71.38$\pm$1.30 & 77.60$\pm$1.81 & 90.23$\pm$0.55  & 88.57$\pm$0.66 \\
    $\,$CDE\textsuperscript{*} & 55.04$\pm$1.73 & 40.08$\pm$1.49 & 68.45$\pm$2.47 & 80.04$\pm$1.75 & 90.05$\pm$0.64 & 87.19$\pm$1.44 \\
    $\,$FLODE & 64.23$\pm$1.84 &  37.16$\pm$1.42 & 73.60$\pm$1.55 & 78.07$\pm$1.62 & 89.02$\pm$0.38 & 86.44$\pm$1.17 \\
    \midrule
    \textbf{Vanilla baseline} \\
$\,$DE-GNN & 63.97$\pm$1.77 & 36.04$\pm$1.08 & 70.99$\pm$2.27  &  76.58$\pm$1.89 & 89.92$\pm$0.59 & 87.03$\pm$1.14 \\
    \midrule
    \textbf{TDE-GNN (ours)} \\
  $\,$TDE-GNN\textsubscript{D} & 70.19$\pm$1.74  & 37.29$\pm$1.19& 77.38$\pm$2.05 & 77.66$\pm$1.91 & 90.28$\pm$0.53  &  87.99$\pm$1.02 \\
  $\,$TDE-GNN\textsubscript{A} & 71.38$\pm$1.93 & 37.02$\pm$1.27 &  78.48$\pm$2.11  & 77.47$\pm$1.82 & 90.08$\pm$0.49 & 87.93$\pm$0.95 \\
    \bottomrule
  \end{tabular}
\end{center}
  \caption{Node classification accuracy ($ \%$). $\uparrow$. \textsuperscript{*} denotes the best result out of several variants. 
  }   \label{table:nodeclassification}

\end{table*}

To demonstrate the efficacy of TDE-GNN, we experiment with two tasks: (i) node classification, and, (ii) spatio-temporal node forecasting, on several benchmarks. 
We provide benchmark details and statistics in Appendix \ref{app:datasets}. The hyperparameters are determined using a  grid search, as discussed in Appendix \ref{app:hyperparameters}. 
Because we propose two possible implementations of the temporal learning mechanism in Section \ref{sec:implementingTime}, we denote the \emph{direct} parameterization variant by TDE-GNN\textsubscript{D} and the \emph{attention} based parameterization variant by TDE-GNN\textsubscript{A}. A detailed description of the TDE-GNN architectures is given in Appendix \ref{app:architectures}. 

We also experiment with a \emph{baseline} model that we call DE-GNN and is implemented according to \Cref{eq:TDE_TimeDisc} with $o=1$ and $c_1=1$, that is, it considers only first-order dynamics, similarly to existing GNNs inspired by DEs. The inclusion of this baseline model to our experiments helps to directly quantify the contribution of our work, TDE-GNN, in addition to the comparison to other GNNs and in particular to other DE-GNNs such as GRAND \citep{chamberlain2021grand}, PDE-GCN \citep{eliasof2021pde}, GRAND++ \citep{thorpe2022grand}, GREAD \citep{choi2022gread}, CDE \citep{convectionGNN}, and FLODE \citep{maskey2023fractional}.
\subsection{Node Classification}
\label{exp:node}
We experiment with homophilic and non-homophilic datasets. The homophilic datasets are Cora \citep{mccallum2000automating}, Citeseer \citep{sen2008collective}, and Pubmed \citep{namata2012query}. The non-homophilic datasets are Chameleon, Squirrel, and Film from \cite{musae}. In all cases, we use the 10 splits from \cite{Pei2020Geom-GCN:}, and report their average accuracy and standard deviation in Table \ref{table:nodeclassification}. 
We consider three types of baselines: (i) `general' GNN architectures, such as GCN \citep{kipf2016semi}, GAT \citep{velickovic2018graph}, GCNII \citep{chen20simple}, Geom-GCN \citep{Pei2020Geom-GCN:}, NSD \citep{bodnar2022neuralsheaf},  GGCN \citep{yan2021two}, H2GCN \citep{zhu2020beyondhomophily_h2gcn}, FAGCN \citep{fagcn2021}, GPRGNN \citep{chien2021adaptive}, GRAFF \citep{di2022graphGRAFF}, LINKX \citep{lim2021large}, and ACMII \citep{luan2022revisiting}. 
(ii) GNNs inspired by differential equations (DEs), including: GRAND \citep{chamberlain2021grand},
PDE-GCN \citep{eliasof2021pde}, GRAND++ \citep{thorpe2022grand}, GREAD \citep{choi2022gread}, CDE \citep{convectionGNN}, and FLODE \citep{maskey2023fractional}. The common theme of those methods is that all consider first-order temporal behavior with $o=1, c_1=1$, except for PDE-GCN, which considers a second-order model, where $o=2, \bfc = [c_1,c_2] = [2,-1]$. In contrast, our TDE-GNN can learn the vector of coefficients $\bfc$ with maximal order $o$, and unless otherwise specified, we set the order to be the number of layers in the network, i.e., $o=L$. The third baseline we consider is (iii) a vanilla version of our TDE-GNN, where $o=1, c_1=1$, and we call this variant DE-GNN.  
Our results in Table \ref{table:nodeclassification} suggest that for homophilic graphs which are known to benefit from diffusion \citep{gasteiger_diffusion_2019}, TDE-GNN performs similarly to other first-order differential equations inspired GNNs, including our baseline DE-GNN, since a diffusion process can be described using a first-order ODE. We find that the significance of learning higher-order dynamics is more pronounced for non-homophilic graphs which may stem from more complex phenomena than homophilic graphs.  For instance, we find that our TDE-GNN\textsubscript{A} achieves an accuracy of 78.48\%, compared to the baseline vanilla, first order DE-GNN with 70.99\% -- a considerable improvement.

\subsection{Spatio-Temporal Forecasting}
\label{exp:spatiotemporal}
\begin{table}[t]
\centering
{
\footnotesize
\setlength{\tabcolsep}{3pt}
\resizebox{1\linewidth}{!}
{
\begin{tabular}{lcccccc}
    \toprule
        \multirow{2}{*}{Dataset}
        & \multicolumn{1}{c}{{Chickenpox}}&  \multicolumn{1}{c}{{PedalMe}} & \multicolumn{1}{c}{Wikipedia} \\
        &Hungary & London & Math\\
\midrule
\textbf{Temporal GNNs} \\
$\,${DCRNN} &    1.124$\pm$0.015            &       1.463$\pm$0.019     &        
{0.679$\pm$0.020}        \\%[0.1cm]
{$\,$GConvGRU} & 1.128$\pm$0.011             &      1.622$\pm$0.032     &          {0.657$\pm$0.015}          \\
$\,${GC-LSTM} &    1.115$\pm$0.014       &      {1.455$\pm$0.023}     &     0.779$\pm$0.023           \\%[0.1cm]
$\,${DyGrAE} & 1.120$\pm$0.021            &      1.455$\pm$0.031     &     0.773$\pm$0.009          \\
$\,${EGCN-O} &    1.124$\pm$0.009       &    1.491$\pm$0.024     &       
0.750$\pm$0.014        \\%[0.1cm]
$\,${A3T-GCN}& {1.114$\pm$0.008}       &     1.469$\pm$0.027  &      0.781$\pm$0.011           \\%[0.1cm]
$\,${T-GCN} & 1.117$\pm$0.011    &    1.479$\pm$0.012     &           0.764$\pm$0.011      \\%[0.1cm]
$\,${MPNN LSTM} & 1.116$\pm$0.023           &       1.485$\pm$0.028      &     0.795$\pm$0.010         \\%[0.1cm]
$\,${AGCRN} & 1.120$\pm$0.010         &      1.469$\pm$0.030   &    0.788$\pm$0.011         \\%[0.1cm]
\midrule
\textbf{Vanilla baseline}\\
$\,$DE-GNN & 0.998$\pm$0.022 & 1.329$\pm$0.041 & 0.714$\pm$0.019\\
\midrule
\textbf{TDE-GNN (ours)} & & & \\
  $\,$TDE-GNN\textsubscript{D} & 0.792$\pm$0.028 & 1.096$\pm$0.057 & 0.614$\pm$0.023 \\
  $\,$TDE-GNN\textsubscript{A} & 0.787$\pm$0.018 & 0.714$\pm$0.051 & 0.565$\pm$0.017 \\
\bottomrule
\end{tabular}
}
}
\caption{The performance of spatio-temporal networks evaluated by the average MSE and standard deviation ($\downarrow$) of 10 experimental repetitions.}\label{tab:predictive_performance}
\end{table}

 We now focus on the applicability of TDE-GNN to spatio-temporal datasets, where the goal is to forecast future node values, given time-series data. We use the Chickenpox-Hungary, PedalMe-London, and Wikipedia-Math datasets from \cite{rozemberczki2021pytorch}.
We use incremental training, mean-squared-error (MSE) loss, and testing procedure from \cite{rozemberczki2021pytorch}.  We report the prediction performance of TDE-GNN, in terms of MSE, in Table \ref{tab:predictive_performance}, and compare it with recent methods like DCRNN \citep{li2018diffusion}, GConv \citep{gconvlstm}, GC-LSTM \citep{gclstm}, DyGrAE \citep{dyggnn, dyngrae_1}, EGCN \cite{evolvegcn}, A3T-GCN \citep{a3tgcn}, T-GCN \citep{tgcn}, MPNN LSTM \citep{panagopoulos2020transfer}, and AGCRN \citep{bai2020adaptive}.

    Because the considered DE-inspired GNNs in the node-classification experiment in Section \ref{exp:node} did not consider spatio-temporal forecasting, and cannot be directly applied to temporal datasets, a direct comparison is not applicable. Instead, we provide the important baseline of a vanilla TDE-GNN that utilizes $o=1, c_1=1$. As previously discussed, this architecture is similar to other first-order DE-inspired GNN models, and we therefore call it DE-GNN. We also note again that including the baseline of DE-GNN allows to \emph{directly} measure the contribution of our TDE-GNN with the learnable temporal in \Cref{eq:TDE_TimeDisc}, and therefore it provides an objective and accurate comparison to our TDE-GNN.
Our results are reported in Table \ref{tab:predictive_performance}, and they show improvement over existing temporal GNN models, as well as the baseline of DE-GNN. This result further highlights the importance of learning higher-order dynamics offered by our TDE-GNN. Also, we see that the attention-based TDE-GNN\textsubscript{A} offers improved performance compared to the directly parameterized TDE-GNN\textsubscript{D}, which can be attributed to the increased complexity of the attention module.

\subsection{Ablation Study}
\label{exp:ablation}
We now study and report two important aspects of our TDE-GNN: the influence of the order $o$, and an analysis of the learned values in $\bfc$.

\begin{figure}[t]
    \centering
    \begin{subfigure}[t]{.49\linewidth}
\includegraphics[width=1\linewidth]{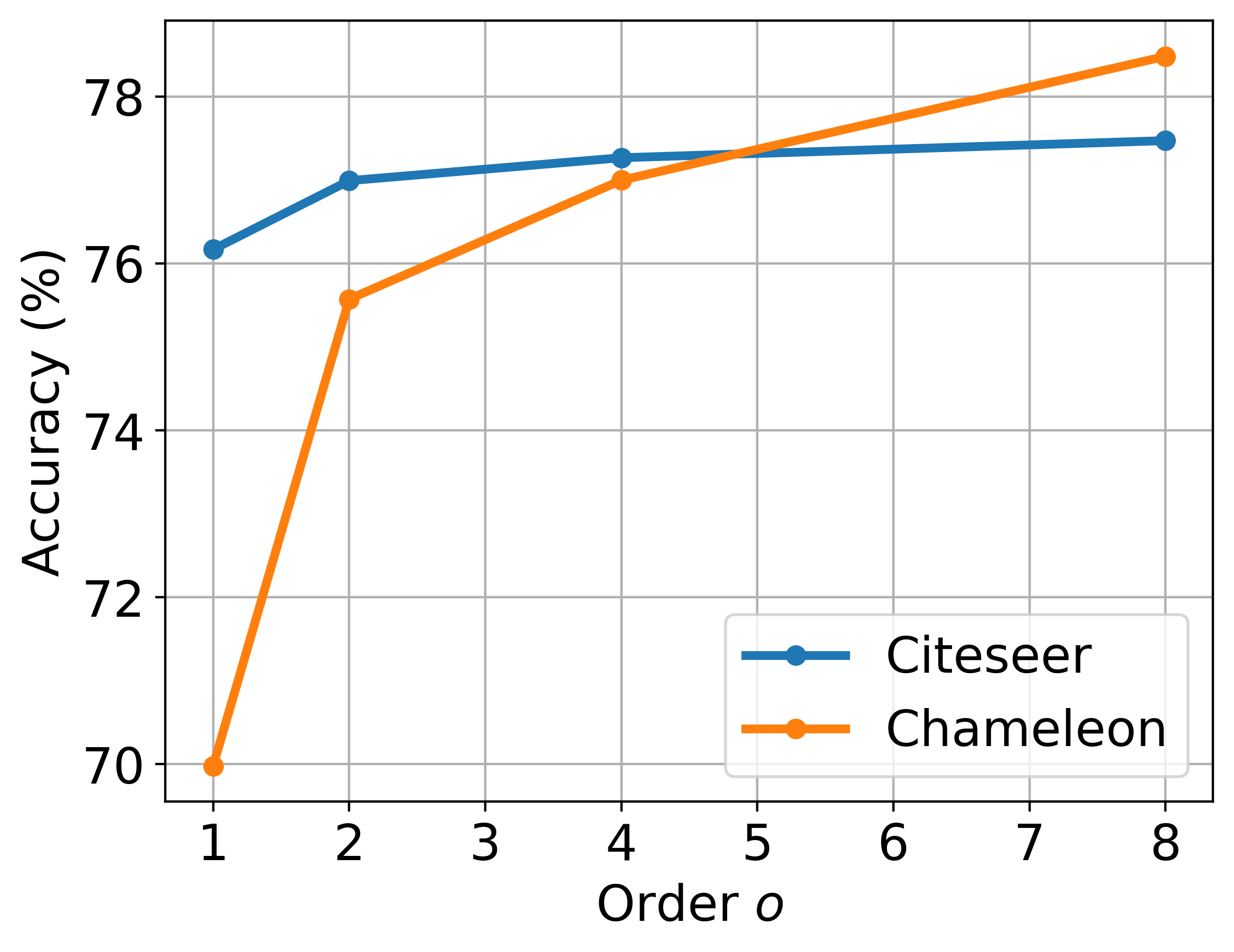}
                    \caption{}
        \label{fig:orderStatic}
\end{subfigure} 
    \begin{subfigure}[t]{.49\linewidth}
    \includegraphics[width=1\linewidth]{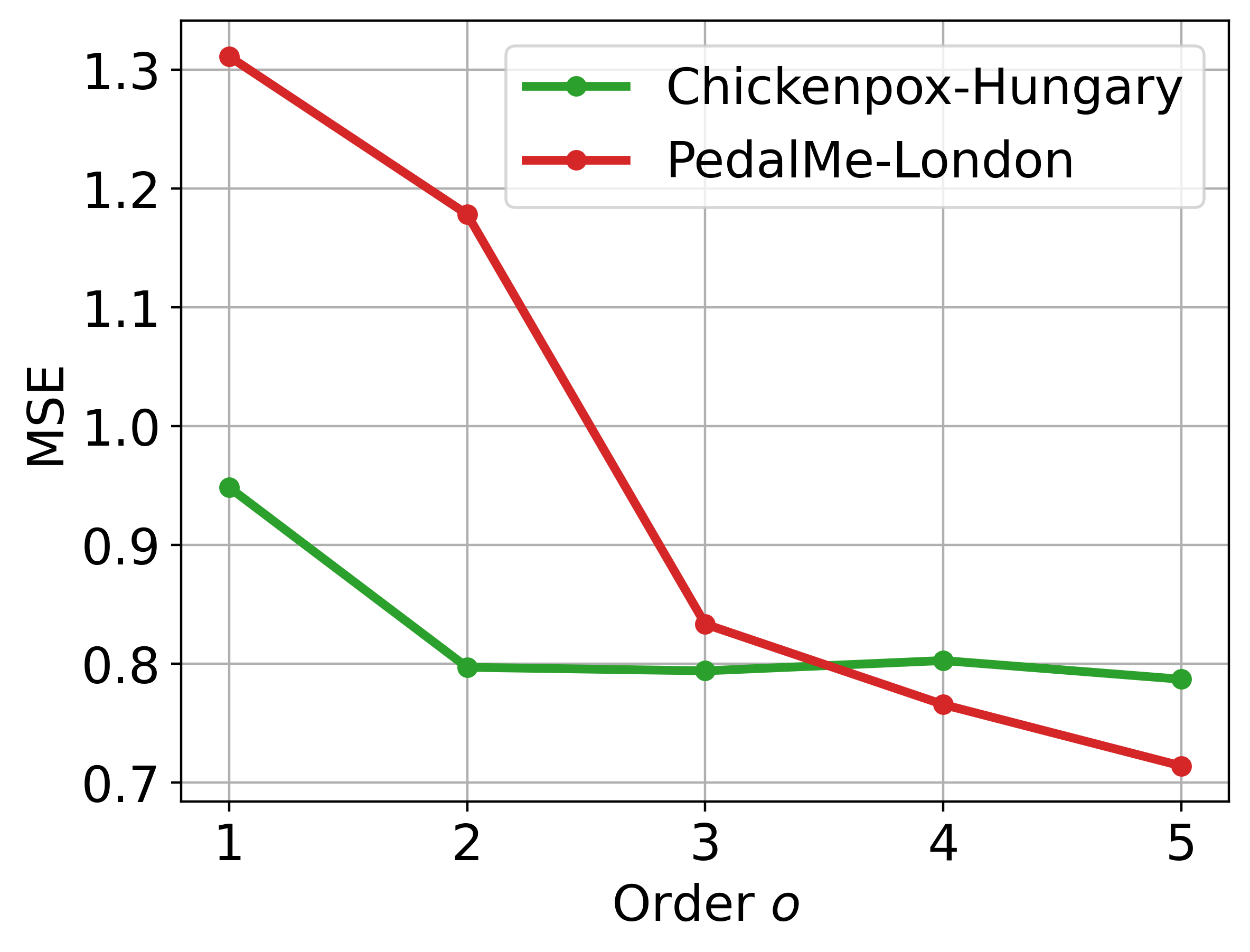}
                        \caption{}
        \label{fig:orderTemporal}
    \end{subfigure}

    \caption{The impact of the model order $o$ on the performance of TDE-GNN.}
    \label{fig:ablationOrder}
\end{figure}

\textbf{The influence of the order $o$.}
As shown in Example \ref{ex:nonlinear_pendulum}, having sufficiently high order can be crucial to modeling complex data. In that example, we have demonstrated this significance via a synthetic task where the order is known. We now supplement this study by reporting the obtained performance as a function of the order $o$ on real-world datasets, where the exact order of the underlying process that generated the data is unknown. To provide a comprehensive study of the impact of $o$, we report the results on both the node classification and spatio-temporal node forecasting tasks considered in this work, on several datasets. The results are reported in Figure \ref{fig:ablationOrder}. Our results indicate, in congruence with Example \ref{ex:nonlinear_pendulum}, that higher-order models can improve performance compared to using a first-order model only, as is common in GNNs. We also note that interestingly, for the Chickenpox-Hungary dataset, we see that a second-order model performs almost as well as a third, fourth, or fifth-order model. While we do not know the exact order of such a real-world dataset, the empirical results may hint at the actual underlying process of the spread of the Chickenpox disease in this dataset being second order.

\textbf{Inspecting the values $\bfc(\mathcal{H}^{(l)}_o)$.} Following Example \ref{ex:nonlinear_pendulum}, where the underlying process of the data is known to be second-order, we now inspect and analyze the obtained coefficients $\bfc$ for a varying order $o \in \{2,3,4,5\}$.\footnote{Recall that an order of $o=1$ cannot describe the second-order data in Example \ref{ex:nonlinear_pendulum}.} As we show in Appendix \ref{app:coefficientsAnalysis}, it is possible to verify that the learned coefficients in Table \ref{tab:coefficients} yield a valid discretization of the second-derivative operator $\frac{\partial^2F}{\partial t^2}$, revealing the true order of the differential equation of the nonlinear pendulum.

\begin{table}[]
    \centering
    \begin{tabular}{|c|c|c|c|c|c|}
    \toprule
    \hline
     & $c_1$ & $c_2$& $c_3$& $c_4$ & $c_5$\\
     \hline
       $o=2$ & 2 & -1 & -- & -- & -- \\
        \hline
        %-0.6 & -0.2 & 1.4 & & \\ 
       $o=3$ & 1.4 & 0.2 & -0.6 & -- & --\\
        \hline
        %-0.4 & -0.25 & 0.675 & 0.975 & \\
        $o=4$ & 0.975 &0.675 &  -0.25 & -0.4 & -- \\
        \hline
        %-0.7604 &  0.0062 &  0.1527 & 1.6799 & -0.0784 \\
     $o=5$ & -0.08 & 1.68 & 0.153 & 0.006 & -0.759  \\
        \hline
        \bottomrule
    \end{tabular}
    \caption{The learned coefficients $\bfc$ with a varying order $o\in\{2,3,4,5\}$ when solving Example \ref{ex:nonlinear_pendulum}.}
    \label{tab:coefficients}
\end{table}

\section{Conclusions}
\label{sec:summary}
In this paper, we studied the temporal domain of GNNs inspired by differential equations. We showed that incorporating higher-order models can be crucial to model data that arises from complex phenomena. This understanding motivated us to develop a novel architecture called TDE-GNN, which utilizes a temporal dynamics learning mechanism, that allows modeling higher-order ODE behaviors in GNNs. Our experimental results show the significance of higher-order GNNs, especially on non-homophilic, and spatio-temporal datasets. Furthermore, the learned temporal coefficients in TDE-GNN allow us to interpret and explain the underlying time-dependent process hidden in the data.

\subsubsection*{Acknowledgements}
ME is a Blavatnik-Cambridge fellow, and is funded by the British Council and the Blavatnik Family Foundation. 
The authors thank Haggai Maron and Beatrice Bevilacqua for the discussions on the attention-based implementation. The authors thank Beatrice Bevilacqua for the discussions and comments on the manuscript.

\bibliographystyle{plainnat}
\bibliography{iclr2021_conference.bib}

%%%%%%%%%%%%%%%%%%%%%%%%%%%%%%%%%%%%%%%%%%%%%%%%%%%%%%%%%%%%

\newpage
\clearpage
\onecolumn
\aistatstitle{Supplementary Materials:  On The Temporal Domain of Differential Equation Inspired Graph Neural Networks
}
\appendix
\section{Proof to Theorem 1}
\label{app:derivations}

We now prove \Cref{thm:stability} from the main paper.

\begin{proof}
Let us write the discrete DE in \Cref{eq:TDE_TimeDisc} explicitly as
\begin{eqnarray}
\label{eq:time_evol}
 \bfF^{(l+1)} =  c_{o} \bfF^{(l-o+1)} + \ldots c_1 \bfF^{(l)} + h s(\bfF^{(l)};G).   
\end{eqnarray}
Similar to the proofs for multistep ODE methods \citep{amr}, to prove the stability of the method, assuming the Jacobians of $s$ have non-positive real part\footnote{If the Jacobian has a positive real part, then the underlying ODE is unstable, and therefore its discretization is also unstable.}, it is sufficient to consider only the temporal term:
$$
\bfF^{(l+1)} = c_{o} \bfF^{(l-o+1)} + \ldots  + c_1 \bfF^{(l)} 
$$
This is a linear, constant-coefficient differential equation, 
and it  must be stable for \Cref{eq:TDE_TimeDisc} to be stable.
Also, as common in proofs of multistep ODE methods \citep{amr}, we start from a solution of the form 
$$ \bfF^{(l)} = \xi^l$$
(meaning $\xi$ to the power of $l$).
Substituting we obtain
$$\xi^{l+1} =    c_o \xi^{l-o+1} + \ldots c_1 \xi^{l} $$
Dividing by $\xi^{l-o+1}$ we obtain the polynomial equation
$$\xi^{o+1} -   c_1 \xi^{o} - \ldots - c_o = 0 $$
This is a polynomial of degree $o+1$ with coefficients $[1, -c_1, \ldots, -c_o]$
where $\sum c_i = 1$.
Let $\rho(c)$ be the roots of the polynomial. 
It is straightforward to see (by substitution) that $1$ is a root of the polynomial.
For the solution to be stable, we need to have that:
\begin{eqnarray*}
&& \bfF^{(n)} = \xi^n \\
&& |\bfF^{(n)}| \le |\bfF^{(n-1)}| \\
&& |\xi^n| \le |\xi^{n-1}| \quad \rightarrow \quad  |\xi| \le 1
\end{eqnarray*}
In multi-step methods for ODEs, this condition is referred to as the root condition. Furthermore, as shown in \cite{amr}, 
since the coefficients $c$ are to be determined (learned, in the case of TDE-GNN), there always exists a set of coefficients such that the root condition is satisfied. 
\end{proof}

\begin{remark}    
While it is difficult to verify the root condition analytically,  it is possible to compute it numerically, thus revealing the order of the process we learn, as we also show in \ref{app:coefficientsAnalysis}.   
\end{remark}

\vfill

\section{Implementation Details}
\label{app:implementation}

\subsection{Implementing the Temporal Term with Attention}
\label{app:implementingTemporal}
We now describe the implementation of TDE-GNN\textsubscript{A}, i.e., the TDE-GNN implemented by an attention mechanism.
Namely, to learn the dynamics between the node features at a current layer $l$ and the previous $o-1$ layers, we utilize a multi-head self-attention mechanism \citep{vaswani2017attention} that assigns scores between the considered layer $l$ and the $o-1$ layers. The difference in our implementation compared to a standard attention module as in \cite{vaswani2017attention} is that we remove the SoftMax normalization step, as discussed in \ref{sec:implementingTime}, and it is required to allow both positive and negative numbers. We denote the attention mechanism by $\rm{MHA}$. The $\rm{MHA}$ computes a score for each pair of layers $(l_i,l_j) \in o \times o$. As an input to the $\rm{MHA}$, we use the history feature tensor as described in \Cref{eq:history}, which is comprised of the stacking of the current and previous (layer-wise) $o-1$ node features. Then, the output of the attention module is given by:
\begin{equation}
    \tilde{\mathcal{S}}_{o}^{(l)} = \label{eq:mha}
    {\rm{MHA}}(\mathcal{H}_o^{(l)}) \in \mathbb{R}^{o\times o}.
\end{equation}
The $(l_i, l_j)$-th entry in $\mathcal{S}_{o}^{(l)}$ represents the connection between the $l_i$-th and $l_j$-th layers. Specifically, the last row of $\mathcal{S}_{o}^{(l)}$ represents the connection between the current layer $l$ and the previous $o-1$ layers. Therefore we define the unnormalized layer coefficients vector as the last row of $\tilde{\mathcal{S}}_{o}^{(l)}$, that can be extracted using the following Python notations:
\begin{equation}
    \label{eq:unnormalizedMHA}
    \tilde{\bfc}(\mathcal{H}_o^{(l)}) = \tilde{\mathcal{S}}_{o}^{(l)}[-1,:] \in \mathbb{R}^{o}
\end{equation}
In order to satisfy condition (i) that demands the weights to sum to 1, described in \ref{sec:implementingTime}, we also add a normalization step, such that the coefficients are defined as:
\begin{equation}
    \label{eq:normalizedMHA}
    {\bfc}(\mathcal{H}_o^{(l)}) = \frac{\tilde{\bfc}(\mathcal{H}_o^{(l)})}{\sum \tilde{\bfc}(\mathcal{H}_o^{(l)})}.
\end{equation}

\subsection{Implementing the Spatial Term}
\label{app:implementingSpace}

The temporal mechanism developed in this paper is generic and can possibly be applied to various DE-GNNs, and is the novelty of our work. However, a GNN inspired by differential equations, and therefore also our TDE-GNN, is not complete without the spatial term that propagates node features across the graph.
As discussed in \Cref{eq:basicODE}, our spatial aggregation function $s$ is based on the combination of a channel mixing operation realized by an MLP and the symmetric normalized graph Laplacian $\bfL = \bfD^{- \frac{1}{2}}(\bfD-\bfA)\bfD^{- \frac{1}{2}}$ where $\bfD$ is the degree matrix, and $\bfA$ is the adjacency matrix of the graph G. Formally, the spatial aggregation is given by:

\begin{equation}
    \label{eq:tdegnn_final2}
    s(\bfF^{(l)}; G) =  \sigma \left( \left(\bfF^{(l)} -  h\bfL\bfF^{(l)}\right)\bfW^{(l)}\right),
\end{equation}
where $\sigma$ is a non-linear activation function, ReLU in our implementation, and $h$ is a positive step size, and $\bfW^{(l)} \in \mathbb{R}^{k \times k}$ are the learnable weights of the MLP.

Therefore, substituting the prescribed temporal in \Cref{eq:tdegnn_final2} into \Cref{eq:TDE_TimeDisc}  leads to our TDE-GNN layer, as follows:
\begin{equation}
    \label{eq:tdegnn_final}
    \bfF^{(l+1)} = \sum_{p=0}^{o-1} c_p(\mathcal{H}_o^{(l)}) \bfF^{(l-p)} + h\sigma \left( \left(\bfF^{(l)} -  h\bfL\bfF^{(l)}\right)\bfW^{(l)}\right).
\end{equation}

\subsection{Architecture details}
\label{app:architectures}
\textbf{Node classification architecture.}
We now elaborate on the TDE-GNN architecture for stationary problems, as used in our node classification experiments. The overall architecture flow is similar to standard GNN architectures for node classification, such as GCN \citep{kipf2016semi} and GCNII \citep{chen20simple}. It is composed of initial embedding layers $e_1,\ldots,e_o$, followed by  $L$ TDE-GNN layers, and a classifier implemented by a linear layer denoted by $e_{out}$. The complete flow of this architecture is described in Algorithm \ref{alg:staticADR}. We train this architecture on node classification datasets by minimizing the cross-entropy loss between the ground-truth node labels $\bfY$ and the predicted node labels $\tilde{\bfY}$.

    \begin{algorithm}[H]
    \caption{TDE-GNN for stationary problems with order $o$.}    \label{alg:staticADR}
    \hspace*{\algorithmicindent} \textbf{Input:} Node features $\bfI^{(0)} \in \mathbb{R}^{n \times k_{in}}$\\
    \hspace*{\algorithmicindent} \textbf{Output:} Predicted node labels $\tilde{\bfY}\in \mathbb{R}^{n \times  k_{out}}$ 
    \begin{algorithmic}[1]
    \Procedure{TDE-GNN}{}
         \State $\bfI^{(0)} \gets {\rm{Dropout}}(\bfI^{(0)}, p)$

        \State $ \bfF^{(-o+1)} = e_1(\bfI^{(0)})$; \ $ \bfF^{(-o+2)} = e_2(\bfI^{(0)})$; \ldots; \ $ \bfF^{(0)} = e_o(\bfI^{(0)})$
        
        \State Initialize history tensor $\mathcal{H}_o^{(0)}$ according to \Cref{eq:history}.

    \For {$l = 0 \ldots L-1$}
       \State $\bfF^{(l)} \gets {\rm{Dropout}}(\bfF^{(l)}, p)$
\State Compute coefficients $\bfc(\mathcal{H}_0^{(l)})$ according to Section \ref{sec:implementingTime}.
\State Update features $\bfF^{(l+1)}$ according to \Cref{eq:tdegnn_final}.
\State Update history tensor $\mathcal{H}_o^{(l+1)}$ according to \Cref{eq:history}.
    \EndFor
         \State $\bfF^{(L)} \gets {\rm{Dropout}}(\bfF^{(L)}, p)$

    \State $\tilde{\bfY} = e_{out}(\bfF^{(L)})$
    \State Return $\tilde{\bfY}$
    \EndProcedure
    \end{algorithmic}
    \end{algorithm}

\textbf{Spatio-Temporal Node Forecasting.}
The typical task in spatio-temporal datasets is to predict future quantities (e.g., driving speed) given several previous time steps (also called frames). Formally, one is given an input tensor $\bfI_{temporal}^{in}= [\bfI^{(0)},\ldots\, \bfI^{(r)}]  \in \mathbb{R}^{n \times 
 r  k_{in}}$, where $r$ is the number of input (observed) time frames, and the goal is to predict $a$ time frames ahead, i.e., the ground-truth is given by ${\bfI}_{temporal}^{gt}=[{\bfI}^{(r+1)},\ldots, {\bfI}^{(r+a)}]\in \mathbb{R}^{n \times ak_{in}}$. This is in contrast to stationary datasets such as Cora \citep{mccallum2000automating}, where input node features $\bfI^{(0)} \in \mathbb{R}^{n \times k_{in}}$ are given, and the goal is to fit to some ground-truth $\bfY \in \mathbb{R}^{n \times k_{out}}$ which can also be of different dimensionality in its output space. In this context, a stationary dataset can be thought of as setting $r = a = 1$ for the non-stationary settings. We show the overall flow of our TDE-GNN architecture for non-stationary problems in Algorithm \ref{alg:temporalADR} \footnote{In Algorithm \ref{alg:temporalADR}, $\oplus$ denotes channel-wise concatenation.}. 

In this architecture, we update the hidden state feature matrix $\bfF^{(l)}_{\rm{state}}$ based on the hidden historical feature matrix $\bfF^{(l)}_{\rm{hist}}$. The reason for this construction is that we want to continue from the current, most recent feature $\bfF^{(l)}_{\rm{state}}$, but also consider the given historical data encoded in $\bfF^{(l)}_{\rm{hist}}$.

Similarly to Attention models \citep{vaswani2017attention}, we incorporate time embedding based on the concatenation of sine and cosine function evaluations with varying frequencies multiplied by the time of the input frames, as input to our TDE-GNN, denoted by ${\bfT_{\rm{emb}}} \in \mathbb{R}^{n \times r k_{temb} }$, where we choose the number of frequencies to be 10, and by the concatenation of both sine and cosine lead to $k_{temb} = 20$. We note that the time embedding is computed in a pre-processing fashion.
To initialize the hidden feature matrices $\bfF^{(0)}_{\rm{state}}, \ \bfF^{(0)}_{\rm{hist}}$, we embed the input data $\bfI_{\rm{temporal}}^{in}$, concatenated with ${\bfT_{\rm{emb}}}$, using two fully connected layers denoted by $e^{\rm{state}}$ and $e^{\rm{hist}}$.

During training, we minimize the mean squared error (MSE) between the ground truth future node quantities and the predicted quantities by TDE-GNN, similar to the training procedure of the rest of the considered methods in Table \ref{tab:predictive_performance}. Specifically, following \cite{rozemberczki2021pytorch}, the goal is to predict the node quantities of the next time frame given 4 previous time frames.

    \begin{algorithm}[H]
    \caption{TDE-GNN for non-stationary problems with order $o$.}    \label{alg:temporalADR}
    \hspace*{\algorithmicindent} \textbf{Input:} Node features $\bfI_{temporal}^{in}=[\bfI^{(0)},\ldots\, \bfI^{(r)}] \in \mathbb{R}^{n \times r k_{in}}$, time embedding ${\bfT_{\rm{emb}}} \in \mathbb{R}^{n \times rk_{temb}}$  \\
    \hspace*{\algorithmicindent} \textbf{Output:} Predicted future node quantities $\tilde{\bfI}_{temporal}^{pred}=[\tilde{\bfI}^{(r+1)},\ldots, \tilde{\bfI}^{(r+a)}]\in \mathbb{R}^{n \times ak_{in}}$ 
    \begin{algorithmic}[1]
    \Procedure{TDE-GNN}{}
        \State $\bfI_{\rm{temporal}}^{in} \gets {\rm{Dropout}}(\bfI_{\rm{temporal}}^{in}, p)$
    \State $\rm{{\bfT_{\rm{emb}}}} \gets  \rm{e^{\rm{time-embed}}}({\bfT_{\rm{emb}}})$
    \State $ \bfF^{(0)}_{\rm{state}} = e^{\rm{state}}(\bfI^{(r)} \oplus {\bfT_{\rm{emb}}})$
        \State $ \bfF^{(0)}_{\rm{hist}} = e^{\rm{hist}}(\bfI_{\rm{temporal}}^{in} \oplus {\bfT_{\rm{emb}}})$
                \State Initialize history tensor $\mathcal{H}_o^{(0)}$ according to \Cref{eq:history}.

    \For {$l = 0 \ldots L-1$}    
    \State $\bfF_{\rm{state}}^{(l)} \gets {\rm{Dropout}}(\bfF_{\rm{state}}^{(l)}, p)$
    \State Compute coefficients $\bfc(\mathcal{H}_0^{(l)})$ according to Section \ref{sec:implementingTime}.
\State Update features $\bfF^{(l+1)}_{state}$ according to \Cref{eq:tdegnn_final}.
\State Update history tensor $\mathcal{H}_o^{(l+1)}$ according to \Cref{eq:history}.
  
    \State $\bfF_{\rm{hist}}^{(l+1)} = e_{l}^{\rm{hist}}(\bfF_{\rm{hist}}^{(l)} \oplus \bfF_{\rm{state}}^{(l+1)} \oplus {\bfT_{\rm{emb}}})$
    \EndFor
    \State $\bfF_{\rm{state}}^{(L)} \gets {\rm{Dropout}}(\bfF_{\rm{state}}^{(L)}, p)$
    
    \State $\tilde{\bfY} = e_{out}^{\rm{state}}(\bfF_{\rm{state}}^{(L)})$
    \State Return $\tilde{\bfI}$
    \EndProcedure
    \end{algorithmic}
    \end{algorithm}

\section{Experimental Details}
\subsection{Pendulum example problem}
\label{app:pendulum}
The pendulum's motion in \Cref{ex:nonlinear_pendulum} is modeled by a time-varying frequency pendulum, such that $$q(F;t) = \sin(\omega(t)F),$$ where $\omega(t)=1 - 0.04\sin(t)$.

\subsection{Benchmarks}
\label{app:datasets}
\textbf{Node classification datasets.}
We report the statistics of the datasets used in our node classification experiments in Table \ref{table:datasets}.
All datasets are publicly available, and appropriate references to the data sources are provided in the main paper.

\begin{table}[h]
  \caption{Node classification datasets statistics.}
  \label{table:datasets}
  \begin{center}
  \begin{tabular}{lccccc}
  \toprule
    Dataset & Classes & Nodes & Edges & Features & Homophily \\
    \midrule
    Squirrel & 5 & 5,201 & 198,493 &  2,089 & 0.22 \\
    Film & 5 & 7,600 & 33,544 & 932 & 0.22  \\
    Chameleon & 5 & 2,277 &  36,101 & 2,325 & 0.23\\
    Citeseer & 6 & 3,327  & 4,732 & 3,703 & 0.80\\
    Pubmed & 3 & 19,717 & 44,338 & 500 & 0.74 \\
   Cora & 7 & 2,708 & 5,429 & 1,433 & 0.81\\
    
    \bottomrule
  \end{tabular}
\end{center}
\end{table}

\textbf{Spatio-temporal forecasting datasets.}
We report the statistics of the datasets used in our spatio-temporal forecasting experiments in Table \ref{tab:desc_discrete}.
All datasets are publicly available, and appropriate references to the data sources are provided in the main paper.

\begin{table}[h]
\centering
\caption{Attributes of the spatio-temporal datasets, and information about the number of time periods ($T$) and spatial units ($|\mathcal{V}|$).}\label{tab:desc_discrete}
{
\begin{tabular}{cccc}
\toprule
Dataset   & Frequency & $T$ & $|\mathcal{V}|$ \\
\midrule
    Chickenpox Hungary & Weekly & 522 & 20 \\
     Pedal Me Deliveries &  Weekly & 36 & 15 \\
 Wikipedia Math & Daily & 731 & 1,068 \\
 \bottomrule
\end{tabular}
}
\end{table}

\subsection{Hyperparameters}
\label{app:hyperparameters}
All hyperparameters were determined by grid search, and the ranges and sampling mechanism distributions are provided in Table \ref{tab:hyperparams}. Also, unless otherwise specified, in all experiments, we use $L=8$ layers, and for node classification datasets, we use $o=L$. For the spatio-temporal datasets, we use $o=r$, i.e., the order is set to be equal to the number of historical data given by the task. 
\begin{table}[h]
\centering
\caption{Hyperparameter ranges}
{
\label{tab:hyperparams}
\begin{tabular}{ccc}
\toprule
Hyperparameter   & Range & Uniform Distribution \\
\midrule
     input/output embedding learning rate &  [1e-4, 1e-1] & log uniform  \\
     temporal term $\bfc$ learning rate &  [1e-4, 1e-1] & log uniform \\
      spatial term learning rate &  [1e-4, 1e-1] & log uniform  \\
       input/output embedding weight decay &  [0, 1e-2] & uniform \\
       temporal term $\bfc$ weight decay & [0, 1e-2] & uniform  \\
    spatial term weight decay & [0, 1e-2] & uniform  \\
      input/output dropout &  [0, 0.9] & uniform \\
      hidden layer dropout & [0, 0.9] & uniform \\
      use BatchNorm & \{ yes / no \} & discrete uniform \\
      step size h &  [1e-3, 1] & uniform \\
      %layers & \{ 2,4,8,16,32,64 \} & discrete uniform \\
      hidden channels $k$ &  \{ 8,16,32,64,128,256 \} & discrete uniform  \\
 \bottomrule
\end{tabular}
}
\end{table}

\subsection{Runtimes}
\label{app:runtimes}
In addition to the complexity discussion in the main paper, we provide the measured runtimes in Table \ref{tab:runtimes}. Learning the temporal order and dynamics requires additional computations compared to the vanilla baseline of DE-GNN, however, it also offers improved performance as we show in our experiments in Section \ref{sec:experiments}. We report the measured training and inferences runtimes, and the number of parameters on the Cora dataset in Table \ref{tab:runtimes}. We measure the runtimes using an Nvidia-RTX3090 with 24GB of memory, which is the same GPU used to conduct our experiments.

\begin{table}[h]
  \caption{Training and inference GPU runtimes (milliseconds), and the number of parameters (thousands).}
  \label{tab:runtimes}
  \begin{center}
  \begin{tabular}{lcccccc}
  \toprule
   \multirow{1}{*}{Metric}  &  DE-GNN ($o=1,c_1=1$) & TDE-GNN\textsubscript{D} ($o=8$) & TDE-GNN\textsubscript{A} ($o=8$)  \\
    \midrule
    Training time &  21.45 & 23.96  & 34.55 \\
    Inference time  &   11.84 & 12.83 &  16.97  \\
    Parameters & 125 & 157 & 174   \\
    \bottomrule
  \end{tabular}
  \end{center}
\end{table}

\subsection{Coefficients analysis}
\label{app:coefficientsAnalysis}
We now conduct an analysis of the learned coefficients $\bfc$ for the pendulum problem in \Cref{ex:nonlinear_pendulum}. For convenience, we present the learned coefficients again, in Table \ref{tab:coefficients_appendix}. Our analysis consists of two parts: (i) stability, and, (ii) consistency.

\begin{table}[h]
    \centering
    \begin{tabular}{|c|c|c|c|c|c|}
    \toprule
    \hline
     & $c_1$ & $c_2$& $c_3$& $c_4$ & $c_5$\\
     \hline
       $o=2$ & 2 & -1 & -- & -- & -- \\
        \hline
       $o=3$ & 1.4 & 0.2 & -0.6 & -- & --\\
        \hline
        $o=4$ & 0.975 &0.675 &  -0.25 & -0.4 & -- \\
        \hline
     $o=5$ & -0.08 & 1.68 & 0.153 & 0.006 & -0.759  \\
        \hline
        \bottomrule
    \end{tabular}
    \caption{The learned coefficients $\bfc$ with a varying order $o\in\{2,3,4,5\}$ when solving Example \ref{ex:nonlinear_pendulum}.}
    \label{tab:coefficients_appendix}
\end{table}

\textbf{Stability analysis.}
Following the derivations in \Cref{thm:stability} proof presented in Appendix \ref{app:derivations}, we examine the root conditions of the learned coefficients. In Table \ref{tab:roots_appendix} we report the absolute values of the characteristic polynomial with the coefficients from Table \ref{tab:coefficients_appendix}.
We note that for orders $2,3$, and $4$, stability is obtained. However, for $o=5$ the coefficients have one unstable mode. This result suggests that one should not use $o=5$ for the pendulum problem in \Cref{ex:nonlinear_pendulum}. However, orders lower than 5 yield stability. We believe that a stable fifth-order model is possible to be learned from the data, however, the incorporation of the root condition to the learning process requires adding constraints to the learning process and therefore is beyond the scope of this work.

\begin{table}[h]
    \centering
    \begin{tabular}{|c|c|c|c|c|c|}
    \toprule
    \hline
     & $|r_1|$ & $|r_2|$& $|r_3|$& $|r_4|$ & $|r_5|$\\
     \hline
       $o=2$ & 1 & 1 & -- & -- & -- \\
        \hline
       $o=3$ & 1 & 0.6 & 1 & -- & --\\
        \hline
        $o=4$ & 1 &1 &  0.629 & 0.629 & -- \\
        \hline
     $o=5$ & 1 & 0.73 & 0.73 & 1.4 & 1  \\
        \hline
        \bottomrule
    \end{tabular}
    \caption{The absolute value of the roots $r_1,\ldots,r_5$ of the characteristic polynomial with coefficients from Table \ref{tab:coefficients_appendix}.}
    \label{tab:roots_appendix}
\end{table}

\textbf{Consistency analysis.} There are two conditions that verify the consistency of the learned coefficients. The first condition requires that the sum of the coefficients $\bfc$ equals to 1, which implies that if the spatial term in \Cref{eq:tdegnn_final}, the future state (node features) is equal to a weighted average of the previous $o-1$ states. This implies that a constant solution can always be achieved. The second condition requires that the application of the coefficients $\bfc$ to a known discrete function yields a consistent approximation to its derivatives of some order. 

\begin{figure}[t]
    \centering    \includegraphics[width=0.7\linewidth]{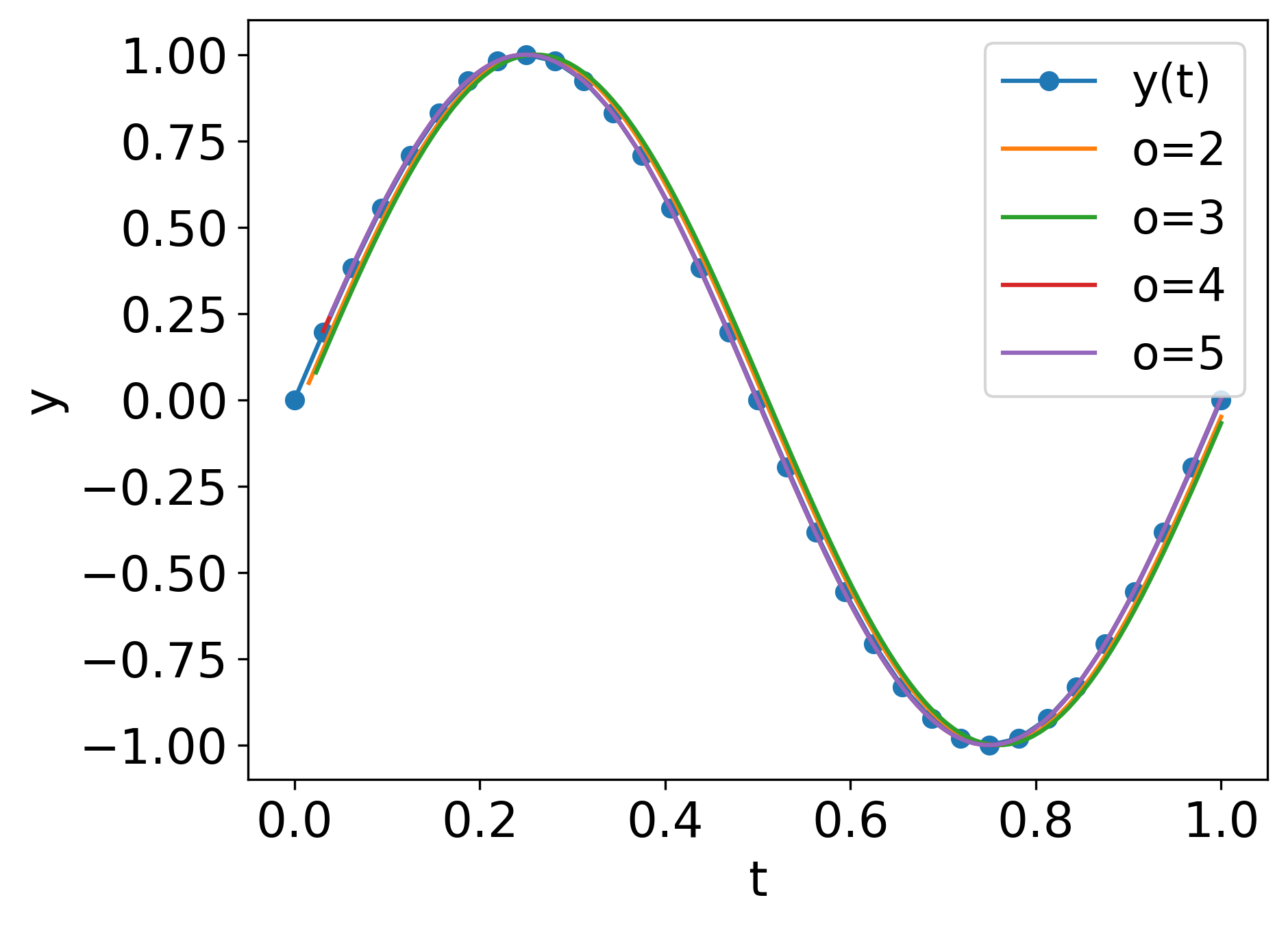} 
    \caption{Examining the consistency of the learned coefficients. The learned coefficients model a second-derivative of the test function $y(t)=\sin(2\pi t)$.}
\label{fig:consistency}
\end{figure}

Note that condition (i) is satisfied by our construction of $\bfc$. The second condition can be verified numerically, as we show now. To this end, we discretize the function $y=\sin(2\pi t)$ in the interval $[0,1]$. We then apply the stencils based on the learned coefficients $\bfc$ as shown in Table \ref{tab:coefficients_appendix}. As can be depicted in Figure \ref{fig:consistency}, the application of the stencils to the discrete function $y(t)$ yields a scaled version of the second derivative of $y(t)$ (which also equals to  $y$, that is, $\frac{\partial^2 y(t)}{\partial t^2} = \beta y(t)$).
This result shows that our TDE-GNN is able to reveal the true order that describes the pendulum's motion, which is a second-order process.  

\end{document}

%% file: math_commands.tex
\usepackage{amsmath,amsfonts,bm}

\def\eqref#1{equation~\ref{#1}}
\def\1{\bm{1}}

\DeclareMathAlphabet{\mathsfit}{\encodingdefault}{\sfdefault}{m}{sl}
\SetMathAlphabet{\mathsfit}{bold}{\encodingdefault}{\sfdefault}{bx}{n}

\newcounter{example}[section]
\newenvironment{example}[1][]{\refstepcounter{example}\par\medskip
   \noindent \textbf{Example~\theexample. #1} \rmfamily}{\medskip}

\newcommand{\bfA}{{\bf A}}

\newcommand{\bfD}{{\bf D}}

\newcommand{\bfF}{{\bf F}}

\newcommand{\bfI}{{\bf I}}

\newcommand{\bfL}{{\bf L}}

\newcommand{\bfT}{{\bf T}}

\newcommand{\bfW}{{\bf W}}

\newcommand{\bfY}{{\bf Y}}

\newcommand{\bfc}{{\bf c}}

%% file: aistats_main.bbl
\begin{thebibliography}{67}
\providecommand{\natexlab}[1]{#1}
\providecommand{\url}[1]{\texttt{#1}}
\expandafter\ifx\csname urlstyle\endcsname\relax
  \providecommand{\doi}[1]{doi: #1}\else
  \providecommand{\doi}{doi: \begingroup \urlstyle{rm}\Url}\fi

\bibitem[Alon and Yahav(2021)]{alon2021oversquashing}
Uri Alon and Eran Yahav.
\newblock On the bottleneck of graph neural networks and its practical
  implications.
\newblock In \emph{International Conference on Learning Representations}, 2021.
\newblock URL \url{https://openreview.net/forum?id=i80OPhOCVH2}.

\bibitem[Ascher et~al.(1995)Ascher, Mattheij, and Russell]{amr}
U.~Ascher, R.~Mattheij, and R.~Russell.
\newblock \emph{Numerical Solution of Boundary Value Problems for Ordinary
  Differential Equations}.
\newblock SIAM, Philadelphia, 1995.

\bibitem[Ascher(2008)]{ascher2008numerical}
Uri~M Ascher.
\newblock \emph{Numerical methods for evolutionary differential equations}.
\newblock SIAM, 2008.

\bibitem[Ascher and Petzold(1998)]{AscherPetzoldODEs}
Uri~M Ascher and Linda~R Petzold.
\newblock \emph{Computer methods for ordinary differential equations and
  differential-algebraic equations}, volume~61.
\newblock Siam, 1998.

\bibitem[Bai et~al.(2020)Bai, Yao, Li, Wang, and Wang]{bai2020adaptive}
Lei Bai, Lina Yao, Can Li, Xianzhi Wang, and Can Wang.
\newblock {Adaptive Graph Convolutional Recurrent Network for Traffic
  Forecasting}.
\newblock \emph{Advances in Neural Information Processing Systems}, 33, 2020.

\bibitem[Bo et~al.(2021)Bo, Wang, Shi, and Shen]{fagcn2021}
Deyu Bo, Xiao Wang, Chuan Shi, and Huawei Shen.
\newblock Beyond low-frequency information in graph convolutional networks.
\newblock In \emph{{AAAI}}. {AAAI} Press, 2021.

\bibitem[Bodnar et~al.(2022)Bodnar, Di~Giovanni, Chamberlain, Li{\`o}, and
  Bronstein]{bodnar2022neuralsheaf}
Cristian Bodnar, Francesco Di~Giovanni, Benjamin Chamberlain, Pietro Li{\`o},
  and Michael Bronstein.
\newblock Neural sheaf diffusion: A topological perspective on heterophily and
  oversmoothing in gnns.
\newblock \emph{Advances in Neural Information Processing Systems},
  35:\penalty0 18527--18541, 2022.

\bibitem[Cai and Wang(2020)]{cai2020note}
Chen Cai and Yusu Wang.
\newblock A note on over-smoothing for graph neural networks.
\newblock \emph{arXiv preprint arXiv:2006.13318}, 2020.

\bibitem[Chamberlain et~al.(2021)Chamberlain, Rowbottom, Gorinova, Bronstein,
  Webb, and Rossi]{chamberlain2021grand}
Ben Chamberlain, James Rowbottom, Maria~I Gorinova, Michael Bronstein, Stefan
  Webb, and Emanuele Rossi.
\newblock Grand: Graph neural diffusion.
\newblock In \emph{International Conference on Machine Learning}, pages
  1407--1418. PMLR, 2021.

\bibitem[Chen et~al.(2018{\natexlab{a}})Chen, Xu, Wu, and Zheng]{gclstm}
Jinyin Chen, Xuanheng Xu, Yangyang Wu, and Haibin Zheng.
\newblock {GC-LSTM: Graph Convolution Embedded LSTM for Dynamic Link
  Prediction}.
\newblock \emph{arXiv preprint arXiv:1812.04206}, 2018{\natexlab{a}}.

\bibitem[Chen et~al.(2020)Chen, Wei, Huang, Ding, and Li]{chen20simple}
Ming Chen, Zhewei Wei, Zengfeng Huang, Bolin Ding, and Yaliang Li.
\newblock Simple and deep graph convolutional networks.
\newblock In Hal~Daumé III and Aarti Singh, editors, \emph{Proceedings of the
  37th International Conference on Machine Learning}, volume 119 of
  \emph{Proceedings of Machine Learning Research}, pages 1725--1735. PMLR,
  13--18 Jul 2020.
\newblock URL \url{http://proceedings.mlr.press/v119/chen20v.html}.

\bibitem[Chen et~al.(2018{\natexlab{b}})Chen, Han, Li, Chen, Xing, Zhao, and
  Li]{chen2018deep}
Weikai Chen, Xiaoguang Han, Guanbin Li, Chao Chen, Jun Xing, Yajie Zhao, and
  Hao Li.
\newblock Deep rbfnet: Point cloud feature learning using radial basis
  functions.
\newblock \emph{arXiv preprint arXiv:1812.04302}, 2018{\natexlab{b}}.

\bibitem[Chien et~al.(2021)Chien, Peng, Li, and Milenkovic]{chien2021adaptive}
Eli Chien, Jianhao Peng, Pan Li, and Olgica Milenkovic.
\newblock Adaptive universal generalized pagerank graph neural network.
\newblock In \emph{International Conference on Learning Representations}, 2021.
\newblock URL \url{https://openreview.net/forum?id=n6jl7fLxrP}.

\bibitem[Cho et~al.(2014)Cho, van Merrienboer, G{\"{u}}l{\c{c}}ehre, Bahdanau,
  Bougares, Schwenk, and Bengio]{GRU}
Kyunghyun Cho, Bart van Merrienboer, {\c{C}}aglar G{\"{u}}l{\c{c}}ehre, Dzmitry
  Bahdanau, Fethi Bougares, Holger Schwenk, and Yoshua Bengio.
\newblock Learning phrase representations using {RNN} encoder-decoder for
  statistical machine translation.
\newblock In Alessandro Moschitti, Bo~Pang, and Walter Daelemans, editors,
  \emph{Proceedings of the 2014 Conference on Empirical Methods in Natural
  Language Processing, {EMNLP} 2014, October 25-29, 2014, Doha, Qatar, {A}
  meeting of SIGDAT, a Special Interest Group of the {ACL}}, pages 1724--1734.
  {ACL}, 2014.
\newblock \doi{10.3115/v1/d14-1179}.
\newblock URL \url{https://doi.org/10.3115/v1/d14-1179}.

\bibitem[Choi et~al.(2023{\natexlab{a}})Choi, Choi, Hwang, Lee, Lee, and
  Park]{choi2023climate}
Hwangyong Choi, Jeongwhan Choi, Jeehyun Hwang, Kookjin Lee, Dongeun Lee, and
  Noseong Park.
\newblock Climate modeling with neural advection--diffusion equation.
\newblock \emph{Knowledge and Information Systems}, 65\penalty0 (6):\penalty0
  2403--2427, 2023{\natexlab{a}}.

\bibitem[Choi et~al.(2023{\natexlab{b}})Choi, Hong, Park, and
  Cho]{choi2022gread}
Jeongwhan Choi, Seoyoung Hong, Noseong Park, and Sung-Bae Cho.
\newblock {GREAD}: Graph neural reaction-diffusion networks.
\newblock In Andreas Krause, Emma Brunskill, Kyunghyun Cho, Barbara Engelhardt,
  Sivan Sabato, and Jonathan Scarlett, editors, \emph{Proceedings of the 40th
  International Conference on Machine Learning}, volume 202 of
  \emph{Proceedings of Machine Learning Research}, pages 5722--5747. PMLR,
  23--29 Jul 2023{\natexlab{b}}.
\newblock URL \url{https://proceedings.mlr.press/v202/choi23a.html}.

\bibitem[Eliasof et~al.(2021)Eliasof, Haber, and Treister]{eliasof2021pde}
Moshe Eliasof, Eldad Haber, and Eran Treister.
\newblock {PDE-GCN}: Novel architectures for graph neural networks motivated by
  partial differential equations.
\newblock \emph{Advances in Neural Information Processing Systems},
  34:\penalty0 3836--3849, 2021.

\bibitem[Eliasof et~al.(2023)Eliasof, Haber, and Treister]{eliasof2023adr}
Moshe Eliasof, Eldad Haber, and Eran Treister.
\newblock Adr-gnn: Advection-diffusion-reaction graph neural networks.
\newblock \emph{arXiv preprint arXiv:2307.16092}, 2023.

\bibitem[Evans(1998)]{EvansPDE}
L.~C. Evans.
\newblock \emph{Partial Differential Equations}.
\newblock American Mathematical Society, San Francisco, 1998.

\bibitem[Gasteiger et~al.(2019)Gasteiger, Wei{\ss}enberger, and
  G{\"u}nnemann]{gasteiger_diffusion_2019}
Johannes Gasteiger, Stefan Wei{\ss}enberger, and Stephan G{\"u}nnemann.
\newblock Diffusion improves graph learning.
\newblock In \emph{Conference on Neural Information Processing Systems
  (NeurIPS)}, 2019.

\bibitem[Giovanni et~al.(2023)Giovanni, Rowbottom, Chamberlain, Markovich, and
  Bronstein]{di2022graphGRAFF}
Francesco~Di Giovanni, James Rowbottom, Benjamin~Paul Chamberlain, Thomas
  Markovich, and Michael~M. Bronstein.
\newblock Understanding convolution on graphs via energies.
\newblock \emph{Transactions on Machine Learning Research}, 2023.
\newblock ISSN 2835-8856.
\newblock URL \url{https://openreview.net/forum?id=v5ew3FPTgb}.

\bibitem[Gravina et~al.(2023)Gravina, Bacciu, and
  Gallicchio]{gravina2023antisymmetric}
Alessio Gravina, Davide Bacciu, and Claudio Gallicchio.
\newblock Anti-symmetric {DGN}: a stable architecture for deep graph networks.
\newblock In \emph{The Eleventh International Conference on Learning
  Representations}, 2023.
\newblock URL \url{https://openreview.net/forum?id=J3Y7cgZOOS}.

\bibitem[Guan et~al.(2022)Guan, Iyer, and Kim]{guan2022dynagraph}
Mingyu Guan, Anand~Padmanabha Iyer, and Taesoo Kim.
\newblock Dynagraph: dynamic graph neural networks at scale.
\newblock In \emph{Proceedings of the 5th ACM SIGMOD Joint International
  Workshop on Graph Data Management Experiences \& Systems (GRADES) and Network
  Data Analytics (NDA)}, pages 1--10, 2022.

\bibitem[Gutteridge et~al.(2023)Gutteridge, Dong, Bronstein, and
  Di~Giovanni]{gutteridge2023drew}
Benjamin Gutteridge, Xiaowen Dong, Michael~M Bronstein, and Francesco
  Di~Giovanni.
\newblock Drew: dynamically rewired message passing with delay.
\newblock In \emph{International Conference on Machine Learning}, pages
  12252--12267. PMLR, 2023.

\bibitem[Haber and Ruthotto(2017)]{haber2017stable}
Eldad Haber and Lars Ruthotto.
\newblock Stable architectures for deep neural networks.
\newblock \emph{Inverse problems}, 34\penalty0 (1):\penalty0 014004, 2017.

\bibitem[He et~al.(2016)He, Zhang, Ren, and Sun]{he2016deep}
Kaiming He, Xiangyu Zhang, Shaoqing Ren, and Jian Sun.
\newblock Deep residual learning for image recognition.
\newblock In \emph{Proceedings of the IEEE Conference on Computer Vision and
  Pattern Recognition}, pages 770--778, 2016.

\bibitem[Hochreiter and Schmidhuber(1997)]{hochreiter1997long}
Sepp Hochreiter and J{\"u}rgen Schmidhuber.
\newblock Long short-term memory.
\newblock \emph{Neural computation}, 9\penalty0 (8):\penalty0 1735--1780, 1997.

\bibitem[Kipf and Welling(2017)]{kipf2016semi}
Thomas~N. Kipf and Max Welling.
\newblock Semi-supervised classification with graph convolutional networks.
\newblock In \emph{International Conference on Learning Representations}, 2017.
\newblock URL \url{https://openreview.net/forum?id=SJU4ayYgl}.

\bibitem[Li et~al.(2018)Li, Yu, Shahabi, and Liu]{li2018diffusion}
Yaguang Li, Rose Yu, Cyrus Shahabi, and Yan Liu.
\newblock {Diffusion Convolutional Recurrent Neural Network: Data-Driven
  Traffic Forecasting}.
\newblock In \emph{International Conference on Learning Representations}, 2018.

\bibitem[Lim et~al.(2021)Lim, Hohne, Li, Huang, Gupta, Bhalerao, and
  Lim]{lim2021large}
Derek Lim, Felix~Matthew Hohne, Xiuyu Li, Sijia~Linda Huang, Vaishnavi Gupta,
  Omkar~Prasad Bhalerao, and Ser-Nam Lim.
\newblock Large scale learning on non-homophilous graphs: New benchmarks and
  strong simple methods.
\newblock In A.~Beygelzimer, Y.~Dauphin, P.~Liang, and J.~Wortman Vaughan,
  editors, \emph{Advances in Neural Information Processing Systems}, 2021.
\newblock URL \url{https://openreview.net/forum?id=DfGu8WwT0d}.

\bibitem[Longa et~al.(2023)Longa, Lachi, Santin, Bianchini, Lepri, Lio,
  Scarselli, and Passerini]{longa2023graph}
Antonio Longa, Veronica Lachi, Gabriele Santin, Monica Bianchini, Bruno Lepri,
  Pietro Lio, Franco Scarselli, and Andrea Passerini.
\newblock Graph neural networks for temporal graphs: State of the art, open
  challenges, and opportunities.
\newblock \emph{arXiv preprint arXiv:2302.01018}, 2023.

\bibitem[Luan et~al.(2022)Luan, Hua, Lu, Zhu, Zhao, Zhang, Chang, and
  Precup]{luan2022revisiting}
Sitao Luan, Chenqing Hua, Qincheng Lu, Jiaqi Zhu, Mingde Zhao, Shuyuan Zhang,
  Xiao-Wen Chang, and Doina Precup.
\newblock Revisiting heterophily for graph neural networks.
\newblock \emph{Conference on Neural Information Processing Systems}, 2022.

\bibitem[Maskey et~al.(2023)Maskey, Paolino, Bacho, and
  Kutyniok]{maskey2023fractional}
Sohir Maskey, Raffaele Paolino, Aras Bacho, and Gitta Kutyniok.
\newblock A fractional graph laplacian approach to oversmoothing.
\newblock \emph{arXiv preprint arXiv:2305.13084}, 2023.

\bibitem[McCallum et~al.(2000)McCallum, Nigam, Rennie, and
  Seymore]{mccallum2000automating}
Andrew~Kachites McCallum, Kamal Nigam, Jason Rennie, and Kristie Seymore.
\newblock Automating the construction of internet portals with machine
  learning.
\newblock \emph{Information Retrieval}, 3\penalty0 (2):\penalty0 127--163,
  2000.

\bibitem[Namata et~al.(2012)Namata, London, Getoor, Huang, and
  Edu]{namata2012query}
Galileo Namata, Ben London, Lise Getoor, Bert Huang, and U~Edu.
\newblock Query-driven active surveying for collective classification.
\newblock In \emph{10th International Workshop on Mining and Learning with
  Graphs}, volume~8, page~1, 2012.

\bibitem[Nt and Maehara(2019)]{nt2019revisiting}
Hoang Nt and Takanori Maehara.
\newblock Revisiting graph neural networks: All we have is low-pass filters.
\newblock \emph{arXiv preprint arXiv:1905.09550}, 2019.

\bibitem[Oono and Suzuki(2020)]{oono2020graph}
Kenta Oono and Taiji Suzuki.
\newblock Graph neural networks exponentially lose expressive power for node
  classification.
\newblock In \emph{International Conference on Learning Representations}, 2020.
\newblock URL \url{https://openreview.net/forum?id=S1ldO2EFPr}.

\bibitem[Panagopoulos et~al.(2021)Panagopoulos, Nikolentzos, and
  Vazirgiannis]{panagopoulos2020transfer}
George Panagopoulos, Giannis Nikolentzos, and Michalis Vazirgiannis.
\newblock {Transfer Graph Neural Networks for Pandemic Forecasting}.
\newblock In \emph{Proceedings of the 35th AAAI Conference on Artificial
  Intelligence}, 2021.

\bibitem[Pareja et~al.(2020)Pareja, Domeniconi, Chen, Ma, Suzumura, Kanezashi,
  Kaler, Schardl, and Leiserson]{evolvegcn}
Aldo Pareja, Giacomo Domeniconi, Jie Chen, Tengfei Ma, Toyotaro Suzumura,
  Hiroki Kanezashi, Tim Kaler, Tao~B Schardl, and Charles~E Leiserson.
\newblock {EvolveGCN: Evolving Graph Convolutional Networks for Dynamic
  Graphs}.
\newblock In \emph{AAAI}, pages 5363--5370, 2020.

\bibitem[Pei et~al.(2020)Pei, Wei, Chang, Lei, and Yang]{Pei2020Geom-GCN:}
Hongbin Pei, Bingzhe Wei, Kevin Chen-Chuan Chang, Yu~Lei, and Bo~Yang.
\newblock Geom-gcn: Geometric graph convolutional networks.
\newblock In \emph{International Conference on Learning Representations}, 2020.
\newblock URL \url{https://openreview.net/forum?id=S1e2agrFvS}.

\bibitem[Pilva and Zareei(2022)]{pilva2022learning}
Pourya Pilva and Ahmad Zareei.
\newblock Learning time-dependent pde solver using message passing graph neural
  networks.
\newblock \emph{arXiv preprint arXiv:2204.07651}, 2022.

\bibitem[Poli et~al.(2019)Poli, Massaroli, Park, Yamashita, Asama, and
  Park]{poli2019graph}
Michael Poli, Stefano Massaroli, Junyoung Park, Atsushi Yamashita, Hajime
  Asama, and Jinkyoo Park.
\newblock Graph neural ordinary differential equations.
\newblock \emph{arXiv preprint arXiv:1911.07532}, 2019.

\bibitem[Rozemberczki et~al.(2021{\natexlab{a}})Rozemberczki, Allen, and
  Sarkar]{musae}
Benedek Rozemberczki, Carl Allen, and Rik Sarkar.
\newblock {Multi-Scale Attributed Node Embedding}.
\newblock \emph{Journal of Complex Networks}, 9\penalty0 (2),
  2021{\natexlab{a}}.

\bibitem[Rozemberczki et~al.(2021{\natexlab{b}})Rozemberczki, Scherer, He,
  Panagopoulos, Riedel, Astefanoaei, Kiss, Beres, L{\'o}pez, Collignon,
  et~al.]{rozemberczki2021pytorch}
Benedek Rozemberczki, Paul Scherer, Yixuan He, George Panagopoulos, Alexander
  Riedel, Maria Astefanoaei, Oliver Kiss, Ferenc Beres, Guzm{\'a}n L{\'o}pez,
  Nicolas Collignon, et~al.
\newblock Pytorch geometric temporal: Spatiotemporal signal processing with
  neural machine learning models.
\newblock In \emph{Proceedings of the 30th ACM International Conference on
  Information \& Knowledge Management}, pages 4564--4573, 2021{\natexlab{b}}.

\bibitem[Rusch et~al.(2022)Rusch, Chamberlain, Rowbottom, Mishra, and
  Bronstein]{rusch2022graph}
T~Konstantin Rusch, Ben Chamberlain, James Rowbottom, Siddhartha Mishra, and
  Michael Bronstein.
\newblock Graph-coupled oscillator networks.
\newblock In \emph{International Conference on Machine Learning}, pages
  18888--18909. PMLR, 2022.

\bibitem[Rusch et~al.(2023)Rusch, Bronstein, and Mishra]{rusch2023survey}
T~Konstantin Rusch, Michael~M Bronstein, and Siddhartha Mishra.
\newblock A survey on oversmoothing in graph neural networks.
\newblock \emph{arXiv preprint arXiv:2303.10993}, 2023.

\bibitem[Ruthotto and Haber(2018)]{RuthottoHaber2018}
Lars Ruthotto and Eldad Haber.
\newblock Deep neural networks motivated by partial differential equations.
\newblock \emph{arXiv preprint arXiv:1804.04272}, 2018.

\bibitem[Sen et~al.(2008)Sen, Namata, Bilgic, Getoor, Galligher, and
  Eliassi-Rad]{sen2008collective}
Prithviraj Sen, Galileo Namata, Mustafa Bilgic, Lise Getoor, Brian Galligher,
  and Tina Eliassi-Rad.
\newblock Collective classification in network data.
\newblock \emph{AI magazine}, 29\penalty0 (3):\penalty0 93--93, 2008.

\bibitem[Seo et~al.(2018)Seo, Defferrard, Vandergheynst, and
  Bresson]{gconvlstm}
Youngjoo Seo, Micha{\"e}l Defferrard, Pierre Vandergheynst, and Xavier Bresson.
\newblock {Structured Sequence Modeling with Graph Convolutional Recurrent
  Networks}.
\newblock In \emph{International Conference on Neural Information Processing},
  pages 362--373. Springer, 2018.

\bibitem[Sun et~al.()Sun, Chen, Xu, Xie, Blum, and
  Venkitasubramaniam]{RDtemporal}
Yue Sun, Chao Chen, Yuesheng Xu, Sihong Xie, Rick~S. Blum, and Parv
  Venkitasubramaniam.
\newblock Reaction-diffusion graph ordinary differential equation networks:
  Traffic-law-informed speed prediction under mismatched data.
\newblock URL \url{https://par.nsf.gov/biblio/10466683}.

\bibitem[Taheri and Berger-Wolf(2019)]{dyngrae_1}
Aynaz Taheri and Tanya Berger-Wolf.
\newblock {Predictive Temporal Embedding of Dynamic Graphs}.
\newblock In \emph{Proceedings of the 2019 IEEE/ACM International Conference on
  Advances in Social Networks Analysis and Mining}, pages 57--64, 2019.

\bibitem[Taheri et~al.(2019)Taheri, Gimpel, and Berger-Wolf]{dyggnn}
Aynaz Taheri, Kevin Gimpel, and Tanya Berger-Wolf.
\newblock Learning to represent the evolution of dynamic graphs with recurrent
  models.
\newblock In \emph{Companion Proceedings of The 2019 World Wide Web
  Conference}, WWW ’19, page 301–307, 2019.

\bibitem[Thorpe et~al.(2022)Thorpe, Nguyen, Xia, Strohmer, Bertozzi, Osher, and
  Wang]{thorpe2022grand}
Matthew Thorpe, Tan~Minh Nguyen, Hedi Xia, Thomas Strohmer, Andrea Bertozzi,
  Stanley Osher, and Bao Wang.
\newblock {GRAND}++: Graph neural diffusion with a source term.
\newblock In \emph{International Conference on Learning Representations}, 2022.
\newblock URL \url{https://openreview.net/forum?id=EMxu-dzvJk}.

\bibitem[Vaswani et~al.(2017)Vaswani, Shazeer, Parmar, Uszkoreit, Jones, Gomez,
  Kaiser, and Polosukhin]{vaswani2017attention}
Ashish Vaswani, Noam Shazeer, Niki Parmar, Jakob Uszkoreit, Llion Jones,
  Aidan~N Gomez, {\L}ukasz Kaiser, and Illia Polosukhin.
\newblock Attention is all you need.
\newblock \emph{Advances in neural information processing systems}, 30, 2017.

\bibitem[Veli{\v{c}}kovi{\'{c}} et~al.(2018)Veli{\v{c}}kovi{\'{c}}, Cucurull,
  Casanova, Romero, Li{\`{o}}, and Bengio]{velickovic2018graph}
Petar Veli{\v{c}}kovi{\'{c}}, Guillem Cucurull, Arantxa Casanova, Adriana
  Romero, Pietro Li{\`{o}}, and Yoshua Bengio.
\newblock {Graph Attention Networks}.
\newblock \emph{International Conference on Learning Representations}, 2018.
\newblock URL \url{https://openreview.net/forum?id=rJXMpikCZ}.

\bibitem[Wang et~al.(2021)Wang, Zhang, Xiao, and Song]{wang2021review}
Jianian Wang, Sheng Zhang, Yanghua Xiao, and Rui Song.
\newblock A review on graph neural network methods in financial applications.
\newblock \emph{arXiv preprint arXiv:2111.15367}, 2021.

\bibitem[Wang et~al.(2022)Wang, Yi, Liu, Wang, and Jin]{wang2022acmp}
Yuelin Wang, Kai Yi, Xinliang Liu, Yu~Guang Wang, and Shi Jin.
\newblock Acmp: Allen-cahn message passing for graph neural networks with
  particle phase transition.
\newblock \emph{arXiv preprint arXiv:2206.05437}, 2022.

\bibitem[Wu et~al.(2020)Wu, Pan, Chen, Long, Zhang, and
  Philip]{wu2020comprehensive}
Zonghan Wu, Shirui Pan, Fengwen Chen, Guodong Long, Chengqi Zhang, and S~Yu
  Philip.
\newblock A comprehensive survey on graph neural networks.
\newblock \emph{IEEE Transactions on Neural Networks and Learning Systems},
  2020.

\bibitem[Xiong et~al.(2023)Xiong, Yang, Jiang, and Ouyang]{diffusionTemporal}
Ni~Xiong, Yan Yang, Yongquan Jiang, and Xiaocao Ouyang.
\newblock Diffusion graph neural ordinary differential equation network for
  traffic prediction.
\newblock In \emph{2023 International Joint Conference on Neural Networks
  (IJCNN)}, pages 1--8, 2023.
\newblock \doi{10.1109/IJCNN54540.2023.10191212}.

\bibitem[Xiong et~al.(2020)Xiong, Ozbay, Jin, and Feng]{xiong2020dynamic}
Xi~Xiong, Kaan Ozbay, Li~Jin, and Chen Feng.
\newblock Dynamic origin--destination matrix prediction with line graph neural
  networks and kalman filter.
\newblock \emph{Transportation Research Record}, 2674\penalty0 (8):\penalty0
  491--503, 2020.

\bibitem[Yan et~al.(2021)Yan, Hashemi, Swersky, Yang, and Koutra]{yan2021two}
Yujun Yan, Milad Hashemi, Kevin Swersky, Yaoqing Yang, and Danai Koutra.
\newblock Two sides of the same coin: Heterophily and oversmoothing in graph
  convolutional neural networks.
\newblock \emph{arXiv preprint arXiv:2102.06462}, 2021.

\bibitem[Zhang et~al.(2019)Zhang, Hao, Wang, de~Silva, and Fu]{zhang2019linked}
Kuangen Zhang, Ming Hao, Jing Wang, Clarence~W de~Silva, and Chenglong Fu.
\newblock Linked dynamic graph cnn: Learning on point cloud via linking
  hierarchical features.
\newblock \emph{arXiv preprint arXiv:1904.10014}, 2019.

\bibitem[Zhao et~al.(2023)Zhao, Kang, Song, She, Wang, and Tay]{convectionGNN}
K.~Zhao, Q.~Kang, Y.~Song, R.~She, S.~Wang, and W.~P. Tay.
\newblock Graph neural convection-diffusion with heterophily.
\newblock In \emph{Proc. International Joint Conference on Artificial
  Intelligence}, Macao, China, Aug 2023.

\bibitem[Zhao et~al.(2019)Zhao, Song, Zhang, Liu, Wang, Lin, Deng, and
  Li]{tgcn}
Ling Zhao, Yujiao Song, Chao Zhang, Yu~Liu, Pu~Wang, Tao Lin, Min Deng, and
  Haifeng Li.
\newblock {T-GCN: A Temporal Graph Convolutional Network for Traffic
  Prediction}.
\newblock \emph{IEEE Transactions on Intelligent Transportation Systems},
  21\penalty0 (9):\penalty0 3848--3858, 2019.

\bibitem[Zhu et~al.(2020{\natexlab{a}})Zhu, Song, Zhao, and Li]{a3tgcn}
Jiawei Zhu, Yujiao Song, Ling Zhao, and Haifeng Li.
\newblock {A3T-GCN: Attention Temporal Graph Convolutional Network for Traffic
  Forecasting}.
\newblock \emph{arXiv preprint arXiv:2006.11583}, 2020{\natexlab{a}}.

\bibitem[Zhu et~al.(2020{\natexlab{b}})Zhu, Yan, Zhao, Heimann, Akoglu, and
  Koutra]{zhu2020beyondhomophily_h2gcn}
Jiong Zhu, Yujun Yan, Lingxiao Zhao, Mark Heimann, Leman Akoglu, and Danai
  Koutra.
\newblock Beyond homophily in graph neural networks: Current limitations and
  effective designs.
\newblock \emph{Advances in Neural Information Processing Systems},
  33:\penalty0 7793--7804, 2020{\natexlab{b}}.

\bibitem[Zhuang et~al.(2020)Zhuang, Dvornek, Li, and
  Duncan]{zhuang2020ordinary}
Juntang Zhuang, Nicha Dvornek, Xiaoxiao Li, and James~S Duncan.
\newblock Ordinary differential equations on graph networks.
\newblock 2020.

\end{thebibliography}
